\documentclass[10pt,journal,compsoc]{IEEEtran}
\ifCLASSINFOpdf
\else
\fi
\usepackage{graphicx}
\usepackage{hyperref}
\usepackage{color}
\usepackage{biblatex} 
\usepackage{multirow}
\usepackage{amsfonts}
\usepackage{amsmath}
\usepackage{bm}
\usepackage{algorithm}
\usepackage{algpseudocode}
\usepackage[table]{xcolor}
\usepackage{comment}

\usepackage[utf8]{inputenx}

\usepackage{todonotes}

\usepackage{pifont}
\newcommand{\xmark}{\ding{55}}%
\newcommand{\cmark}{\ding{51}}%

\newcommand{\revone}[1]{{\color{black} #1}} 
\newcommand{\revtwo}[1]{{\color{black} #1}} 
\newcommand{\freespace}[1]{{\color{black} #1}} 

\begin{document}
%
\title{A survey on efficient vision transformers: algorithms, techniques, and performance benchmarking}
%
%
%
%

\author{Lorenzo~Papa,~\IEEEmembership{Student Member,~IEEE,}
        Paolo~Russo,
        Irene~Amerini,~\IEEEmembership{Member,~IEEE,}
        and~Luping~Zhou,~\IEEEmembership{Senior~Member,~IEEE}
\IEEEcompsocitemizethanks{
\IEEEcompsocthanksitem L. Papa is the corresponding author. The work has been developed during the visiting research period at The University of Sydney, Australia, NSW, 2006.
\IEEEcompsocthanksitem L. Papa, P. Russo and I. Amerini are with the  Department of Computer, Control and Management Engineering, Sapienza University of Rome, Italy, IT, 00185.
E-mail: [papa, paolo.russo, amerini]@diag.uniroma1.it
\IEEEcompsocthanksitem L. Papa and L. Zhou are with the School of Electrical and Information Engineering, Faculty of Engineering, The University of Sydney, Australia, NSW, 2006. E-mail: [lorenzo.papa, luping.zhou]@sydney.edu.au}
\thanks{Manuscript received April 19, 2005; revised August 26, 2015.}
\thanks{L. Zhou would like to acknowledge the The USyd-Fudan BISA Flagship Research Program (2023), I. Amerini would like to acknowledge SERICS (PE00000014) under the MUR National Recovery and Resilience Plan funded by the European Union - NextGenerationEU, Sapienza University of Rome project 2022–2024 “EV2” (003\_009\_22), and L. Papa would like to acknowledge project 2022–2023 “RobFastMDE”.}}

%
%

\markboth{Journal of \LaTeX\ Class Files,~Vol.~14, No.~8, August~2015}%
{Shell \MakeLowercase{\textit{et al.}}: Bare Advanced Demo of IEEEtran.cls for IEEE Computer Society Journals}
%



\IEEEtitleabstractindextext{%
\begin{abstract}
Vision Transformer (ViT) architectures are becoming increasingly popular and widely employed to tackle computer vision applications. 
Their main feature is the capacity to extract global information through the self-attention mechanism, outperforming earlier convolutional neural networks.
However, ViT deployment and performance have grown steadily with their size, number of trainable parameters, and operations.
Furthermore, self-attention's computational and memory cost quadratically increases with the image resolution. 
Generally speaking, it is challenging to employ these architectures in real-world applications due to many hardware and environmental restrictions, such as processing and computational capabilities. 
Therefore, this survey investigates the most efficient methodologies to ensure sub-optimal estimation performances.
More in detail, four efficient categories will be analyzed: compact architecture, pruning, knowledge distillation, and quantization strategies.
Moreover,  a new metric called \textit{Efficient Error Rate} has been introduced in order to normalize and compare models' features that affect hardware devices at inference time, such as the number of parameters, bits, FLOPs, and model size.
Summarizing, this paper firstly mathematically defines the strategies used to make Vision Transformer efficient, describes and discusses state-of-the-art methodologies, and analyzes their performances over different application scenarios.
Toward the end of this paper, we also discuss open challenges and promising research directions.
\end{abstract}

\begin{IEEEkeywords}
Computer vision, computational efficiency, vision transformer
\end{IEEEkeywords}}

\maketitle

\IEEEdisplaynontitleabstractindextext

%
\IEEEpeerreviewmaketitle

\section{Introduction}
\IEEEPARstart{A}{rtificial} intelligence (AI) solutions based on deep learning (DL) infrastructures are becoming increasingly popular in a variety of everyday life and industrial application scenarios, such as chat-bots and perception systems~\cite{devlin2018bert, liu2019roberta, ding2022davit, liu2022swin, zhai2022scaling}.
Those tasks are usually based on neural language processing (NLP) and computer vision (CV) solutions, specifically focusing on text and image analysis.
\revtwo{Although such algorithms have usually been developed through convolutional neural networks (CNN) models, i.e., architectures consisting of convolutional operations used to extract information at different scales, recently, new families of neural networks such as \cite{vaswani2017attention, fu2023monarch, tolstikhin2021mlp} have been proposed.
These methodologies exhibit exceptional performance in AI applications and consistently push their limits. 
They have in common the self-attention mechanism, which simultaneously extracts specific information from each input data, including text prompts and pixels, while also considering their inter-relationships for a sensible improvement of the simple translation-invariant property of the convolution operator.}
Generally speaking, the attention mechanism has been introduced by Bahdanau et al.~\cite{bahdanau2014neural} (2014) in order to address the bottleneck problem that arises in encoder-decoder architectures to flexibly translate the most relevant information of the encoded input sequences into the decoder part.
In NLP tasks, this issue is especially true for lengthy and/or sophisticated sequences, while in CV-dense prediction applications such as semantic segmentation and depth estimation, this technique will lead to superior reconstruction capabilities of the model's outputs, as shown in~\cite{chen2022vitadapter, yuan2022neural, papa2023meter}. 
Furthermore, the attention mechanism is then reformulated by Vaswani et al.~\cite{vaswani2017attention} (2017) in order to extract intrinsic features while showing the self-attention potential in order to capture long-range dependencies.

Since their first developments, transformer models have usually been applied to NLP scenarios such as language-to-language translation tasks~\cite{luong2015stanford, bojar2016findings}.
\revtwo{Subsequently, inspired by the remarkable performances achieved by the global receptive field of transformer architectures in CV tasks, Dai et al.~\cite{dai2017deformable} (2017) introduce the deformable convolutions in order to overcome the fixed geometric structures of CNN modules.
Similarly, Wang et al.~\cite{wang2018non} (2018) present non-local operation techniques for capturing long-range dependencies, showing that non-local neural networks outperform well-known 2D and 3D convolutional architectures for video classification tasks.}
Moreover, Zhang et al.~\cite{zhang2020dynamic} (2020) propose to model long-rate dependencies via graph convolutional neural network. 
The authors introduce a dynamic graph message passing network (DGMN) which is able to achieve substantial improvements with respect to dynamic convolution operations while using fewer MAC operations.
These results are achieved thanks to the larger receptive field and less-redundant information captured by graph nodes.
Despite these solutions, which demonstrate the ability to improve CNN models, Dosovitskiy et al.~\cite{dosovitskiy2020image} (2021) propose the first solution in order to apply the self-attention to bi-dimensional signals, namely vision transformers (ViT).
The authors design a general backbone where the input image is divided into fixed-size patches in order to apply the standard self-attention. 
\revtwo{This preliminary work exhibits the capabilities and performances of transformers also in the CV research field.
However, one of the major challenges for ViT is the computational cost of their key element, i.e., the self-attention itself; this behavior is particularly true for high-resolution and dense prediction tasks~\cite{wang2023image, conde2022swin2sr, yuan2022neural, papa2023meter}.}
In particular, self-attention's computational and memory cost increases quadratically with the image resolution. 
Moreover, the Softmax operation computed in the attention block makes these structures computationally demanding for edge and low-resource devices.
These demanding hardware requirements provide significant hurdles for ViT models to infer on resource-constrained devices such as embedded devices and autonomous systems. 
Furthermore, to provide a high-quality user experience, real-time computer vision systems incorporating transformer-based models must fulfill low latency requirements.
Consequently, due to the growing development of novel ViT architectures with improved estimation performances (which comes at the expense of elevated computational costs) and the need for resource-constrained devices in real-world AI tasks, this survey aims to give an in-depth analysis of the most recent solutions in order to design \textbf{efficient ViT models}.

Generally speaking, previously proposed surveys by Han et al.~\cite{han2022survey} (2022), Tay et al.~\cite{tay2022efficient} (2022), and Khan et al.~\cite{khan2022transformers} (2022) mainly focus on a general overview of transformers models from both NLP and CV tasks and their various application scenarios.
Specifically, those earlier studies exclusively rely on small subsections, future investigations, or a limited amount of analyzed methodologies related to the efficient ViT strategies.
Furthermore, several works have been recently conducted (2023) in response to the rising need for more accurate models capable of performing tasks in real-world settings, resulting in significant progress in the research field.
\revtwo{In contrast to previous surveys, Zhuang et al.~\cite{zhuang2023efftrainoftrans} (2023) recently presented a systematic overview specifically focused on the efficient training of Transformers models.
Consequently, through the examination of several current efforts, and in order to give a more comprehensive overview of efficiency methodologies in ViT, we observed the necessity of a survey entirely focused on the efficiency-architectural domain.}
Precisely, as briefly mentioned also in~\cite{han2022survey}, we categorize efficient deep learning algorithms for ViT structures into four categories by adopting the following reported methodology:
\begin{itemize}
    \item \textit{Compact architecture} (CA) - analyzes solutions specifically developed to reduce the computational cost of self-attention in order to guarantee the ViT global understanding of the input features while reducing (often by attention linearization) the computational cost of such architectures.
    \item \textit{Pruning} (P) - focuses on strategies designed to reduce the number of neurons and connections of ViT models in order to maintain high accuracy while avoiding the model over parametrization and reducing the number of computed operations (multiplications). 
    \item \textit{Knowledge distillation} (KD) - analyzes learning strategies that aim to improve the performance of shallow (student) models by sharing and compressing the knowledge from deeper ones (teacher).
    \item \textit{Quantization} (Q) - technologies that aim to reduce data type, from floating point to integer, and precision, from 32-bit to a lower bit-rate, of ViT's weights and activation functions in order to obtain lightweight and memory-efficient models. 
\end{itemize}

Moreover, the main contributions of this work are summarized as follows:
\begin{itemize}
    \item We present a comprehensive review of efficient ViT methodologies emphasizing the mathematical aspects, analyzing the proposed strategies, and comparing their performances on well-known benchmark datasets.  
    \item We surveyed the milestones of efficient ViT strategy (up to 2023) for the four selected efficient categories (CA, P, KD, and Q).
    \item We introduce a novel evaluation metric named Efficient Error Rate (EER), which is able to consider all the parameters that can affect a general device at inference time, such as the number of parameters, bits, FLOPs, and model size, in order to compare the analyzed methodologies over well-known benchmark datasets fairly. We also identify the best strategy that better balances EER and estimation capabilities (accuracy).
    \item We finally provide some useful insight for future development and promising research directions. 
\end{itemize}

The rest of the survey is organized as follows:
Section~\ref{sec:general_background} reviews some general and mathematical concepts of ViT and efficient strategies, 
Section~\ref{sec:eff_ViT} describes state-of-the-art efficient deep learning solutions proposed in recent years.
Finally, Section~\ref{sec:results_and_applications} compares the estimation performances achieved by the previously introduced methods when applied to tackle different CV tasks, and Section~\ref{sec:conclusion_and_discussions} discusses ViT challenges and promising research directions.

\section{Background}
\label{sec:general_background}
This section aims to define preliminary notions and mathematical formulations for the four categories of the analyzed efficient techniques in order to highlight better the novelties introduced in the state-of-the-art researches reported in Section \ref{sec:eff_ViT}.
Precisely, following the four subset categorization of the survey, Section~\ref{sec:background_attention_attention}, Section~\ref{sec:pruning_background}, Section~\ref{sec:KD_background}, and Section~\ref{sec:quantization_background}, respectively formalize vision transformers, pruning, knowledge distillation, and quantization strategies.

\subsection{Background on Vision Transformers}
\label{sec:background_attention_attention}
\revtwo{The development of transformer architectures has been mainly due to their ability to capture long-range dependencies and incorporate more information with respect to fully CNN models when trained on very large datasets~\cite{truvron2022eccv}. 
Those architectures are composed of two key elements: patch embedding and feature extraction module computed via cascaded self-attention blocks. 
Moreover, classification heads or other structures, such as decoders, are stacked on top of the encoding part in order to perform the desired task.
For instance, in the case of a general classification task, the first element is used to process the input RGB image before the feature extraction takes place in order to obtain a sequence of embedded data.
Subsequently, each embedded sequence is fed into a transformer encoder composed of multiple self-attention blocks in order to extract low-level features and complex relationships between different elements of each input sequence.
Finally, a multi-layer perception (MLP) head composed of two feed-forward functions is used to compute the output probabilities.
We report in the following paragraphs a more detailed mathematical formulation of the path embedding and self-attention mechanism.}

\vspace{1.0em}
\noindent
\textbf{Patch embedding:} The first key element of a transformer structure is the creation of patch embedding; in order to obtain it, the input feature maps are usually processed as follows: at first, the input image $x\in R^{H\times W\times C}$ is divided into $N$ patches. Similar to a word sequence in NLP tasks, a patch is a pixel matrix containing part of the input image, i.e., a subset of the input data.  Each patch is then flattened in order to obtain a sequence of $n$ entities and multiplied with a trainable embedding tensor which learns to linearly project each flat patch to dimension $d$;
this results in $n$ embedded patches of shape $1 \times d$, lets generally denote them as $N\in R^{n\times d}$.
Subsequently, a trainable positional embedding is added to the sequence of projections in order to add the spatial representation of each patch within the image space, let's define the overall output embedding as $z$.

\begin{figure}[t]
    \centering
    \includegraphics[width=0.9\linewidth]{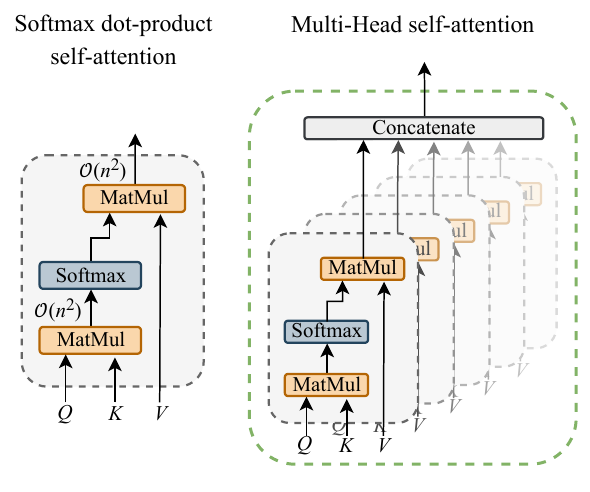}
    \caption{Graphical overview of the vanilla self-attention and the multi-head self-attention blocks. The $\mathcal{O}(n^2)$ in the softmax dot-product self-attention highlights the quadratic cost of each operation.}
    \label{fig:softamx_dot_product_self_attention_CA}
\end{figure}

\vspace{1.0em}
\noindent
\textbf{Self-attention:} As a following step, given the patch embedding sequence $z_n$, the self-attention mechanism learns how to gather one token $t_i\in z_n$ with the others ($t_j$ with $j\neq i$ and $i,j\in d$) into the sequence.
This solution leads to global information extraction from the input features, which improves the fixed receptive field of well-known convolution operations. 
Usually, the transformer structure is based on a multi-head self-attention (MSA) mechanism, which is composed of several single self-attention layers running in parallel; we report a graphical overview of the just introduced operations in Figure~\ref{fig:softamx_dot_product_self_attention_CA}.

Therefore, the self-attention module can be mathematically defined as follows.
Given an input vector, this operation first computes three matrices: the query $Q$, the key $K$, and the value $V$, respectively, with equal sizes ($d_q = d_k = d_v$). 
Subsequently, the operation translates the obtained scores into probabilities, computing the $softmax$ function.
Consequently, the Vanilla self-attention~\cite{vaswani2017attention}, also known as softmax dot-product self-attention operation, can be formally defined with the attention matrix $A \in R^{n\times d_v}$ as reported in Equation~\ref{eq:attention} where the output matrix $A$ updates each component of a sequence by aggregating global information from the complete input sequence.
\begin{equation}
    \label{eq:attention}
    A(Q,K,V) = Softmax \bigg ( \frac{Q\cdot K^T}{\sqrt{d_k}} \bigg ) \cdot V
\end{equation}

However, from a computational complexity aspect, the time and memory cost for this operation increases quadratically with the number of patches $n$ within an image, i.e., $\mathcal{O}(n^2)$.
This cost is due to two dot product operations (reported as MatMul in Figure~\ref{fig:softamx_dot_product_self_attention_CA}), i.e., the computation of $\frac{QK^T}{\sqrt{d_k}}$, which takes $\mathcal{O}(n^2d_k)$, and the second one between the $softmax$ probability and $V$ which takes $\mathcal{O}(n^2d_v)$. 
\revtwo{Then, self-attention modules are concatenated together into an MSA in order to extract information from different areas at the same time. 
Finally, the inputs and output of the MSA module are normalized ($Norm$) and passed into a feed-forward network (FNN) consisting of two feed-forward layers interleaved by a nonlinear activation function (usually a GeLU).} 
Then, by defining as $X$ the input features of the transformer module, it is possible to formulate its output $X_{out}$ as expressed in Equation~\ref{eq:out_transf}:
\begin{equation}\begin{split}
    \label{eq:out_transf}
    & X_{MSA} = Norm(MSA(X,X)) + X \\
    & X_{out} = Norm(FNN(X_{MSA})) + X_{MSA}
\end{split}\end{equation}

\vspace{1.0em}
\noindent
\revtwo{\textbf{Cross-attention:} Based on a similar computation to the self-attention operation, the cross-attention mechanism is also widely employed in vision transformer applications after the patch embedding calculation. 
More in detail, the cross-attention leverages the use of two separate embedding sequences, computing the matrix $Q$ from the first stream and $K$ and $V$ from the second one, i.e., the patch embedding operation is performed twice for each embedded data. 
This solution allows the conditioning of deep learning models with extra information, such as text prompts or other images.}

\subsection{Background on Pruning}
\label{sec:pruning_background}
The neural network pruning strategy, introduced by Janowsky et al.~\cite{janowsky1989pruning} (1989) and Karnin et al.~\cite{karnin1990simple} (1990), is inspired by the synaptic pruning in the human brain.
During this procedure, only a subset of the connections, i.e., axons and dendrites, will remain connected between the childhood and puberty phases. 
In neural networks, this solution does not require interfering with training procedure and loss functions; precisely, the pruning strategy is computed after the training phase in order to reduce the number of neurons and connections of the network; in fact, it is also known as post-training pruning.
This procedure is performed by setting to zero some neuron's weights, usually, the ones with the lower estimation contribute (saliency), i.e., neurons that have minimal impact in the final prediction.
Generally speaking, this strategy will result in lighter models with respect to the original ones, which are usually called “sparse” models due to the spare (i.e., with zeros) matrix obtained after pruning.
More in detail, under the assumption that the used framework is able to take advantage of sparse computation on CPU/GPU devices, this strategy will reduce the number of multiplications between layers at inference time, saving hardware computation and improving the model's inference frequency.
Moreover, by defining with $D$ the dataset, $f(x, W)$ the trained model, where $x$ is the input data ($x\in D$), and $W$ its weights, we can formalize the pruning algorithm as reported in Algorithm~\ref{alg:pruning}.
\begin{algorithm}
    \caption{Pruning strategy}
    \label{alg:pruning}
    \hspace*{\algorithmicindent} \textbf{Input:} $f(\cdot, \cdot)$, trainable neural newton \\
    \hspace*{\algorithmicindent} \textbf{\hspace{0.9cm}} $D$, set training of data \\
    \hspace*{\algorithmicindent} \textbf{\hspace{0.9cm}} $P_{0/1}$, matrix of pruned features
    \begin{algorithmic}[1]
        \State $f(\cdot, W_0) \leftarrow weights\ initialization$
        \State $f(x\in D, W') \leftarrow train\ until\ convergence$
        \State $f(x \in d \subseteq D, P_{0/1} \odot W') \leftarrow pruning$ 
    \end{algorithmic}
\end{algorithm}

Precisely, after the training phase, the pruning strategy is obtained by multiplying the trained weights $W'$ by a diagonal binary matrix  $P_{0/1} \in \{0, 1\}^{|W'|}$ which is composed of a set of pruned features $p_{0/1}$ with $p_{0/1} \in \{0, 1\}^\mathbb{R}$. 
Moreover, to select the latter features it is commonly used a subset $d \subseteq D$ in order to compute the weights' importance score. 
Consequently, after the pruning operation, only a subset of the trained weights $W_{P_{0/1}} \in W'$ will not be set to zero.

However, due to the reduction of the number of neurons and connections inside the original model, this strategy usually leads to a reduction of the generalization capabilities with respect to the original model. 
Therefore, in order to define the best trade-off between zeroed weights (fast inference) and estimation performances, the developed pruning strategies will mainly differ on how to identify the weights that can be zeroed with the minor accuracy reduction, i.e., in how to define the $p_{0/1}-$vector optimally.
In order to give a general overview, pruning techniques usually focus on three sets of studies: (1) \textit{structure} and \textit{unstructured} methods, i.e., strategies that focus on the entire set of weights or on specific ones. 
(2) \textit{score computation}, which are the possible ways to compute the pruning vector and the identification of the number of network's weights that can be pruned.
(3) \textit{training phase}, which focuses on the training-pruning (or fine-tuning) strategy employed at the training phase.

\subsection{Background on Knowledge Distillation}
\label{sec:KD_background}
\begin{figure}[h]
    \centering
    \includegraphics[width=\linewidth]{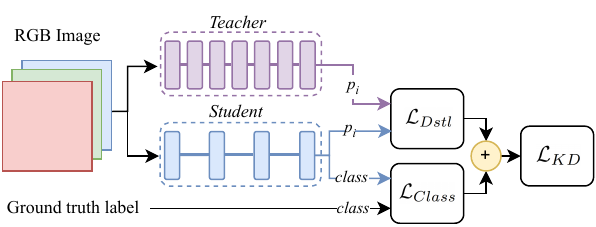}
    \caption{Graphical representation of the vanilla KD learning strategy. Please refer to Section~\ref{sec:KD_background} for the used notation.}
    \label{fig:vanilla_KD}
\end{figure}

The knowledge distillation (KD) strategy has been introduced by Hilton et al.~\cite{hinton2015distilling} (2015) in order to reduce the computational requirements of deep neural networks with respect to other solutions like ensemble learning and the mixture of experts.
This choice is particularly due to the difficult deployment of large deep neural network models on edge devices with limited memory and computational capacity. 
The basic idea behind this KD strategy is to transfer the generalization and estimation abilities from a deep (pretrained) model to a shallower and lightweight one through class probabilities.
As commonly defined, we identify the deeper model as the \textit{teacher} and the shallower as the \textit{student}.
Therefore, the objective of the learning strategy is to train the student model in order to match the class probabilities ($p_i$) produced by the teacher; the authors define these values as soft targets.
Consequently, we can define the distillation loss ($\mathcal{L}_{Dstl}$) as reported in Equation~\ref{eq:kd_distill_loss} where $p_i^t$ and $p_i^s$ are respectively the soft labels of the teacher and the student and $CE$ the cross-entropy loss function.
\begin{equation}
    \label{eq:kd_distill_loss}
    \mathcal{L}_{Dstl} = CE(p_i^t, p_i^s)
\end{equation}
The described procedure allows the smaller model to minimize the distance from the teacher's output distribution, i.e., learning from information that is not provided by the ground-truth labels, and consequently, closely mimic the behavior of the pretrained large teacher.
Moreover, by adding to the overall optimization problem the cross-entropy loss computed between the hard labels (classification vector) of the student and the ground truth labels, which we define as ($\mathcal{L}_{Class}$), we can formulate the loss function of the vanilla KD learning strategy ($\mathcal{L}_{KD}$) as reported in Equation~\ref{eq:ce_loss_kd}.\\
\begin{equation}
    \label{eq:ce_loss_kd}
    \mathcal{L}_{KD} = \mathcal{L}_{Class} + \mathcal{L}_{Dstl}
\end{equation}
Furthermore, to give a better understanding of the newly introduced learning strategy, we show in Figure~\ref{fig:vanilla_KD} its block diagram representation.

\begin{table*}[!t]
    \centering
    \footnotesize
    \caption{Summary of all the analyzed efficient vision transformer models. 
    We organize the modes following their categorization, i.e., compact architecture design (CA), pruning methods (P), knowledge distillation (KD), and quantization (Q). We also underline methods that combine (+) different strategies.}
    \begin{tabular}{l | l | c c }
         Category & Model / Paper & General overview & Code \\
         \hline
         \multirow{24}*{CA} & \multirow{2}*{PVT (Wang et al., 2021)} & Propose the SRA attention module which has & \multirow{2}*{\cmark} \\ 
          & & a computational cost $R_i^2$ time lower than MSA & \\ 
         \cline{2-4}
         & \multirow{2}*{Swin Transformer (Liu et al., 2021)} & 
         Propose a shifted window attention (non overlap- & \multirow{2}*{\cmark} \\ 
         & &  ping patches) against vanilla sliding window &  \\ 
         \cline{2-4}
         & \multirow{2}*{SOFT (Lu et al., 2021)} & Introduce a softmax-free attention module & \multirow{2}*{\cmark} \\ 
          & & obtained via low-rank decomposition strategy & \\ 
         \cline{2-4}
         & \multirow{2}*{PoolFormer (Yu et al., 2022)} & Replace the attention with a non-parametric & \multirow{2}*{\cmark} \\ 
          &  &  pooling operation to reduce the computation & \\ 
         \cline{2-4}
         & \multirow{2}*{PVTv2 (Wang et al., 2022)} & Introduce a Linear SRA attention module which & \multirow{2}*{\cmark} \\ 
         &  & leverage the pooling spatial reduction capabilities & \\ 
         \cline{2-4}         
         & \multirow{2}*{MViTv2 (Li et al., 2022)} & Limited computational complexity  & \multirow{2}*{\cmark} \\ 
          & & with residual pooled attention & \\ 
         \cline{2-4}
         & \multirow{2}*{SimA (Koohpayegani and Pirsiavash, 2022)} &  In order to avoid $exp(\cdot)$ operations, authors & \multirow{2}*{\cmark} \\ 
          & & propose a softmax-free attention solution & \\ 
         \cline{2-4}
         & \multirow{2}*{Flowformer (Huang et al., 2022)} & Introduce an attention module & \multirow{2}*{\cmark} \\ 
          & & inspired by flow neural networks & \\ 
         \cline{2-4}
         & \multirow{2}*{\revone{Hydra Attention (Bolya et al., 2022)}} & \revone{Linear attention based on decomposable kernel}  & \multirow{2}*{\revone{\cmark}} \\ 
          & & \revone{strategies and an high amount of attention heads} &  \\ 
         \cline{2-4}
         & \multirow{2}*{Ortho (Huang et al., 2022)} &  Reduce the computational complexity of MSA & \multirow{2}*{\xmark} \\ 
          & & orthogonalizing the tokens within local regions  &  \\ 
         \cline{2-4}
         & \multirow{2}*{Castling-ViT (You et al., 2023)} &  Propose a fully-linear ViT architecture & \multirow{2}*{\cmark} \\ 
         & & thought a linear-angular attention module  &  \\ 
         \cline{2-4}
         & \multirow{2}*{\revone{EfficientViT (Cai et al., 2023)}} &  \revone{Propose a fully-linear ViT architecture designed} & \multirow{2}*{\revone{\cmark}} \\ 
         & &  \revone{for high-resolution dense prediction tasks}  &  \\ 
         \hline
         \multirow{2}*{CA + Q} & \multirow{2}*{EcoFormer (Liu et al., 2023)} & Propose to kernelizing the attention module via & \multirow{2}*{\cmark} \\ 
          & & hash functions to reduce its computational cost & \\ 
         \hline
         \multirow{10}*{P} & \multirow{2}*{VTP (Zhu et al., 2021)} & Propose a strategy to prune & \multirow{2}*{\xmark} \\ 
         &  & attention modules & \\ 
         \cline{2-4}
         & \multirow{2}*{DPS-ViT (Tang et al., 2022)} & Propose a path slimming algorithm to & \multirow{2}*{\xmark} \\ 
         & & identify and discard redundant patches & \\ 
         \cline{2-4}
         & \multirow{2}*{WDPruning (Yu et al., 2022)} & Introduce a solution to reduce both & \multirow{2}*{\xmark} \\ 
         & & weight and depth of a ViT & \\ 
         \cline{2-4}
         & \multirow{2}*{NViT (Yang et al., 2023)} & Propose an Hessian-based global & \multirow{2}*{\cmark} \\ 
         & & structured pruning algorithm & \\ 
         \cline{2-4}
         & \multirow{2}*{X-Pruner (Yu and Xiang, 2023)} & Design a layer-wise pruning algorithm & \multirow{2}*{\cmark} \\ 
         & & based on eXplainable AI masks  & \\ 
         \hline
         \multirow{2}*{P + KD}& \multirow{2}*{DynamicViT (Rao et al., 2023)} & Leverage to progressively and dynamically &  \multirow{2}*{\cmark} \\ 
         & & prune less informative ViT tokens & \\ 
         \hline
         \multirow{14}*{KD} & \multirow{2}*{DeiT (Touvron et al. 2021)} & Introduce a distillation token into the attention & \multirow{2}*{\cmark} \\ 
         & & module to mimic teacher's hard labels & \\ 
         \cline{2-4}
         & \multirow{2}*{Monifold Distillation (Hao et al., 2022)} & Introduce a loss function based on patch-level  & \multirow{2}*{\xmark} \\ 
         & & information present in transformers modules & \\ 
         \cline{2-4}
         & \multirow{2}*{TinyViT (Wu et al., 2022)} & Introduce a new family of small ViT which & \multirow{2}*{\cmark} \\ 
         & & is trained via memory efficient KD strategy & \\ 
         \cline{2-4}
         & \multirow{2}*{DearKD (Chen et al., 2022)} & Introduce a distillation framework & \multirow{2}*{\xmark} \\ 
         & & for limited/data-free training & \\ 
         \cline{2-4}
         & \multirow{2}*{CivT (Ren et al., 2022)} & Distillation strategy based on multiple teachers & \multirow{2}*{\cmark} \\ 
         & & which extract features from different perspectives & \\ 
         \cline{2-4}
         & \multirow{2}*{MiniViT (Zhang et al., 2022)} & Introduce a weights multiplexing strategy & \multirow{2}*{\cmark} \\ 
         & & applied to ViT architectures & \\ 
         \cline{2-4}
         & \multirow{2}*{SMKD (Lin et al., 2023)} & Propose a patch-masking approach for & \multirow{2}*{\cmark} \\ 
         & & ViT applied to few-shot learning tasks & \\ 
         \hline
         \multirow{10}*{Q} & \multirow{2}*{- (Liu et al., 2021)} & Introduce a mixed-precision weights strategy & \multirow{2}*{\xmark} \\ 
         & & formulated as an optimization problem & \\ 
         \cline{2-4}
         & \multirow{2}*{PTQ4ViT (Yuan et al., 2022)} & Twin uniform strategy, which splits negative & \multirow{2}*{\cmark} \\ 
         & & and positive weight's values into two bit-ranges & \\ 
         \cline{2-4}
         & \multirow{2}*{APQ-ViT (Ding et al., 2022)} & Popose a Matthew-effect preserving  & \multirow{2}*{\xmark} \\ 
         & & scheme for ultra-low bit quantization & \\ 
         \cline{2-4}
         & \multirow{2}*{Auto-ViT-Acc (Lit et al., 2022)} & Introduce a ViT quantization strategy  & \multirow{2}*{\xmark} \\ 
         & & specifically designed for FPGA devices & \\ 
         \cline{2-4}
         & \multirow{2}*{NoisyQuant (Liu et al., 2023)} & Improve 
         previous quantization methods by adding & \multirow{2}*{\xmark} \\ 
         & &  a fixed noisy factor to the date-distribution &  \\ 
         \hline
         \multirow{2}*{Q + P + KD}& \multirow{2}*{GPU-SQ-ViT (Yu et al., 2023)} & Introduce a GPU friendly framework  & \multirow{2}*{\xmark} \\ 
         & & based on a mixed KD strategy & \\ 
         \hline
    \end{tabular}
    \label{tab:summary_survey}
\end{table*}

\subsection{Background on Quantization}
\label{sec:quantization_background}

The last compression approach that we review in this work is the quantization procedure. 
The quantization has the objective of reducing the neural network parameters (weight and activation values) from floating point precision data types, i.e., usually 32-bit, to a lower bit representation, such as 8/6/4/2-bit precision and/or to a different data type, i.e., integers.
Thus, quantized models will significantly save storage memory and speed up the inference process.
However, due to the strong compression, the model could suffer severe accuracy drops or even instabilities.

\begin{figure}[h]
    \centering
    \includegraphics[width=0.8\linewidth]{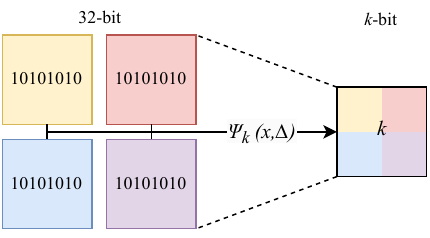}
    \caption{Graphical representation of a general quantization procedure; the floating-point 32-bit data is compressed based on the quantization function $\Psi_k (x, \Delta)$ to a \textit{k}-bit representation.}
    \label{fig:quantization_overview}
\end{figure}

The quantization function $\Psi(\cdot, \cdot)$ and the quantization intervals $\Delta$ (also known as scaling factors, i.e., the way to represent the data with the available bit-width) are the two main critical features of this approach, which should be carefully chosen.
To give a general overview, by defying with $x$ a floating-point value and with $k$ the number of available bits, we report in Equation~\ref{eq:quant_range_eq_split} the most popular function, namely the uniform quantization function, where the data range is equally split.
\begin{equation}
    \label{eq:quant_range_eq_split}
    \Psi_k (x, \Delta) = Clamp \bigg( Round \bigg( \frac{x}{\Delta} \bigg), -2^{k-1}, 2^{k-1} -1 \bigg)
\end{equation}
Moreover, a graphical representation of the quantization procedure from the commonly used 32-bit representation to a general \textit{k}-bit compression is reported in Figure~\ref{fig:quantization_overview}.

Precisely, there exist two types of quantization methods: the quantization-aware training (QAT) and the post-training quantization (PTQ).
The first strategy interleaves training and quantization procedures, i.e., finetuning the quantized model at low precision, while the second is computed after the training phase without the need for re-training or finetuning.
However, for ViT architecture, the QAT strategy is highly costly due to the re-training phases; differently, PTQ could be a preferable solution, enabling a fast quantization and deployment, needing a few samples (also unlabelled) to calibrate the quantization procedure.

\begin{figure*}[t]
    \centering
    \includegraphics[width=\linewidth]{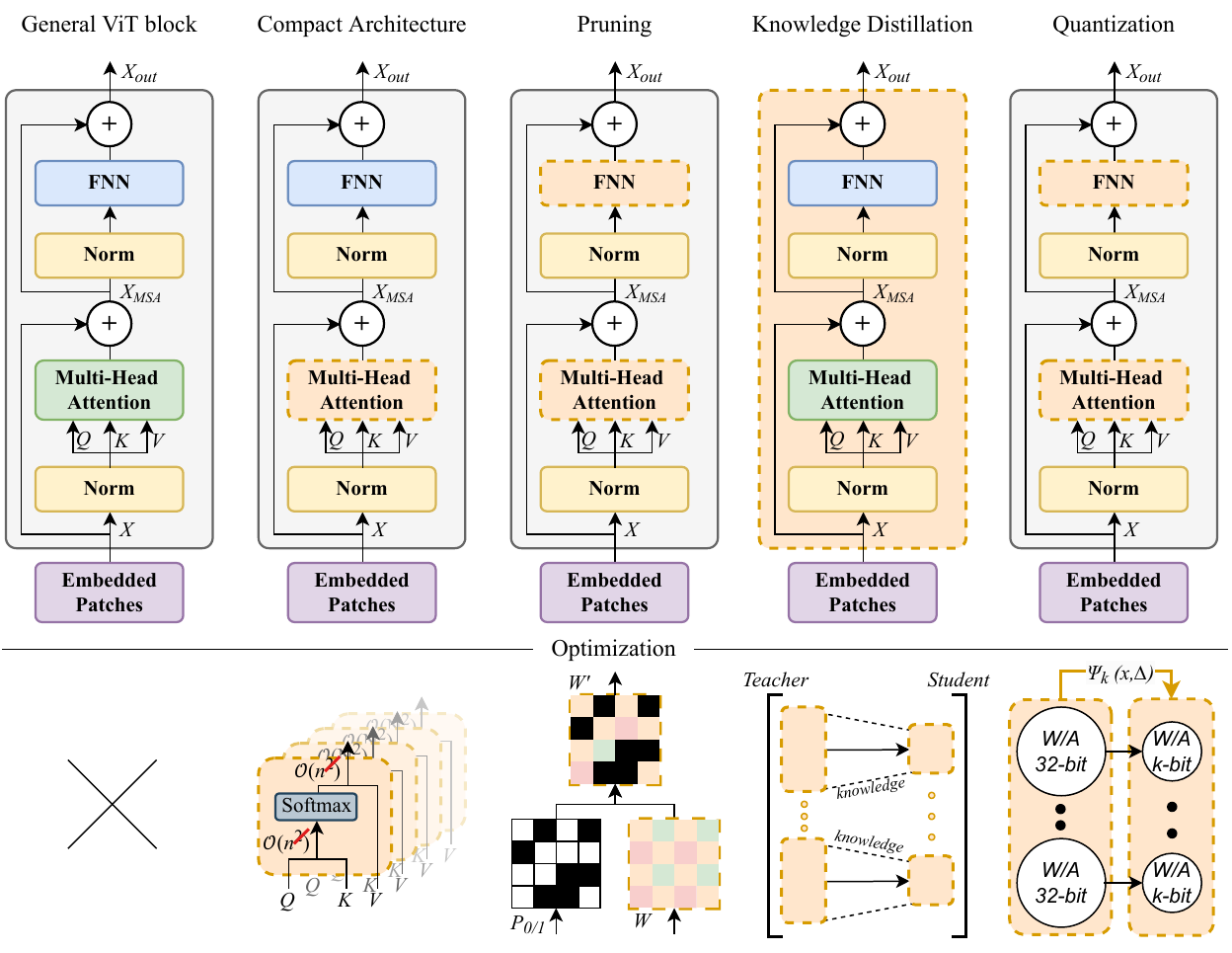}
    \caption{Graphical representation of efficient ViT techniques and their optimization effects. 
    The fundamental element on which the VITs are built is shown on top, where the dashed orange block highlights the component on which each optimization technique mainly focuses. A graphical depiction of how the optimizations influence the block of interest is also provided at the bottom of the image.
    Please refer to Section~\ref{sec:general_background} for the reported variables and their description.}
    \label{fig:compression_overview}
\end{figure*}

\section{Efficient Vision Transformer}
\label{sec:eff_ViT}
This section reviews architectures and strategies proposed in recent years in order to design efficient ViT models.
Precisely, following the four subset categorization of the survey, Section~\ref{sec:compact_arch}, Section~\ref{sec:pruning_decomp}, Section~\ref{sec:knowledge_distill}, and Section~\ref{sec:quantization}, respectively analyzes the compact ViT, pruning, knowledge distillation and quantization strategies specifically designed for ViT models.
We report in Figure \ref{fig:compression_overview} a graphical overview of the compression technique behaviors and a summary of the reviewed models in Table \ref{tab:summary_survey}.

\subsection{Compact architecture design}
\label{sec:compact_arch}
Based on the mathematical basis introduced in Section~\ref{sec:background_attention_attention}, in this paragraph, we review multiple researches focused on architectural-like optimizations, i.e., solutions that are designed to reduce the computational cost of ViT by reducing or linearizing the cost of the self-attention module.
In order to give a better understanding of how these technologies have developed, the proposed solutions are reviewed to address the task according to their release date. 
Such a straightforward solution will emphasize a gradual reduction of the self-attention's computational cost. In particular, we will point out that earlier research (2021) will reduce the quadratic computational cost without attaining a fully-linear design, as will be proposed in more recent studies (2023).

A preliminary work has been conducted by Wang et al.~\cite{wang2021pyramid} (2021), who propose Pyramid Vision Transformer (PVT), a versatile convolution-free (pyramidal) backbone for dense prediction tasks. The ViT architecture uses a progressive shrinking strategy based on a pyramidal structure in order to handle high-resolution feature maps and reduce their computational cost.
The paper introduces and substitutes the standard multi-head attention (MSA) with spatial-reduction attention (SRA), lowering the computational/memory complexity of attention operation.
Moreover, the SRA reduces the resource consumption of ViTs while making the PVT model flexible to learning multi-scale and high-resolution features, i.e., working with high-resolution images.
Then, the SRA is $R_i^2$\footnote{$R_i$ is the reduction ratio of the attention layers in Stage $i$.} times lower than the standard MSA; therefore, the model can handle high-resolution input features with lower computational/memory requirements.
\revtwo{Furthermore, by defining with $SR(\cdot)$ the operation for reducing the spatial dimension of the input sequence, it is possible to formally define the single attention module $A_{SRA}$ of the SRA\footnote{SRA is then obtained as MSA concatenating $h$ attention modules} in Equation~\ref{eq:pvt_att}.}
\begin{equation}
    \label{eq:pvt_att}
    A_{SRA}(Q, K, V) = A(Q, SR(K), SR(V))
\end{equation}

During the same year, Liu et al.~\cite{liu2021swin} (2021) propose Swin Transformer, an architecture able to achieve a linear computation complexity to input image size.
This result is achieved by computing a local multi-head self-attention operation, i.e., only within each (non-overlapping) window (W-MSA).
This choice avoids a global self-attention operation as proposed in Dosovitskiy et al., taking advantage of smaller feature windows obtained aggregating neighboring patches of the input feature maps.
Moreover, the authors introduce a \textit{shifted window} (SW) operation computed between consecutive self-attention layers.
In regards to the standard \textit{sliding window} operation, the proposed solution, combined with the self-attention mechanism, namely shifted W-MSA (SW-MSA), leads to notable latency improvements in real-world applications facilitating memory access in hardware.

Differently from previous studies, Lu et al.~\cite{lu2021soft} (2021) propose SOFT, a softmax-free transformer model.
The authors identify the high computational cost of vision transformer architectures in the softmax operation (i.e., $exp(\cdot)$) computed in the self-attention layer, therefore proposing to approximate the operation via low-rank decomposition.
Therefore, the final formulation of the softmax-free self-attention ($A_{SOFT}$) is formulated as follows, the authors generate $\tilde{Q}$ and $\tilde{K}$ with a convolution and average pooling operations from the query $Q$ and key $K$ respectively.
\begin{equation}
    A_{SOFT} = exp \big( Q \ominus \tilde{K} \big) \cdot \Big( exp \big (\tilde{Q}\ominus \tilde{K} \big) \Big)^\dag \cdot exp \big( \tilde{Q} \ominus K \big)
\end{equation}

The following year, Yu et al.~\cite{yu2022metaformer} (2022) propose a study on the impact of the token mixer in vision transformers; the authors abstract the general structure into the so-called MetaFormer structure. 
Furthermore, by replacing commonly used token mixers, such as the self-attention operation, with a non-parametric pooling ($Pool$) operator, Yu et al. designed PoolFormer.
\revtwo{Therefore, given the input features $X$, the MSA module of PoolFormer ($X_{PF_{MSA}}$) can by formulated as reported in the following equation.}
\begin{equation}
    \label{eq:poolformer_att}
    X_{PF_{MSA}} = Pool(Norm(X) + X)  
\end{equation}
Based on the reported formulation, thanks to the pooling operation, the proposed model is able to achieve a linear computational complexity to the token's sequence length without the addition of any learnable parameter.

Li et al.~\cite{li2022mvitv2} (2022) propose MViTv2, an improvement of the previously proposed MViT backbone by Haoqi Fan et al.~\cite{fan2021multiscale} (2021). 
More in detail, MViTv2 improves the pooling self-attention introduced in MViT~\cite{fan2021multiscale}, with a residual pooling connection, in order to keep a limited computational complexity and reduced memory requirements of the attention block while increasing the information flow and facilitate the training process.
This operation is performed by adding a pooled query tensor $Pool(Q)$ to the output sequence inside the MViT attention module.
\revtwo{Therefore, by defining the pooling self-attention $A_{MViT}$ as reported in Equation~\ref{eq:poolself_att}, its improved version proposed $A_{MViTv2}$ can be formulated as reported in Equation~\ref{eq:poolself_att_imp}.}
\begin{equation}
    A_{MViT} = A \big( Pool(Q), Pool(K), Pool(V) \big)
    \label{eq:poolself_att}
\end{equation}
\begin{equation}
    A_{MViTv2} = A_{MViT} + Pool(Q)
    \label{eq:poolself_att_imp}
\end{equation}

Moreover, inspired by prior works that employ the pooling operation in order to reduce the computational cost of self-attention, Wang et al.~\cite{wang2022pvt} (2022) provide an enhanced version of PVT, namely PVTv2.
The innovative architecture is designed on a novel linear spatial-reduction attention (Linear SRA) module, which uses an average pooling to reduce the spatial dimension before the attention operation rather than the convolutional one utilized in the previous SRA module.
In addition, the authors also introduce an overlapping patch embedding to prevent losing some of the image's local continuity.

Motivated by the issue highlighted by Lu et al.~\cite{lu2021soft} in 2021, Koohpayegani and Pirsiavash~\cite{koohpayegani2022sima} (2022) provide an enhanced strategy to manage the exponential operation computed by the softmax layer.
The authors introduce SimA, a softmax-free attention block, in which the softmax layer is replaced with a $l_1$-normalization operation in order to normalize the query $(\hat{Q}=||Q||_1)$ and key ($\hat{K}=||K||_1$) matrices.
This solution modifies the computation of vanilla self-attention into a standard sequence of multiplications between matrices.
More in detail, similarly to Lu et al.~\cite{lu2021soft}, this choice is due to the $exp(\cdot)$ operation computed in the softmax layer, which, in a transformer architecture, does most of the computation when the input feature map is large.
However, differently from Lu et al.~\cite{lu2021soft}, SimA is not based on low-rank decomposition; instead, it is based on an adaptation method in order to linearize the transformer's computational cost at test time.
Moreover, the authors demonstrate the effectiveness of the proposed solution applied to state-of-the-art models such as DeiT, XCiT, and CvT, achieving on-par accuracy with respect to standard self-attention layers.
Although the overall computational complexity has been reduced (it is more efficient when $n>>d$), the overall cost of the proposed model is still quadratic with respect to the number of input tokens ($\mathcal{O}(n^2d)$) or the dimension of each token ($\mathcal{O}(nd^2)$).
\revtwo{However, SimA is able to dynamically choose the smallest computational complexity based on the number of input tokens linearizing the overall computational cost ($\mathcal{O}(nd)$); the SimA attention module $A_{SimA}$ can be formulated as reported in Equation~\ref{eq:att_simA}.}
\begin{equation}
    \label{eq:att_simA}
    A_{SimA} =  \begin{cases}
    \hat{Q}(\hat{K^T}V) & \textrm{\ if $n > d$} \\ 
    (\hat{Q}\hat{K^T})V & \textrm{\ otherwise}
    \end{cases}
\end{equation}
Summarizing, SimA is able to adapt its computation at the test time in order to achieve a linear computation on the number of tokens or on the number of channels while being more efficient on edge and low-power devices thanks to the softmax-free solution, i.e., lack of $exp(.)$ operations. 

Differently from all the previous studies, Huang et al.~\cite{huang2022flowformer} (2022) propose Flowformer, a transformer architecture based on \textit{Flow-attention} modules.
The optimized attention layer is inspired by flow networks, architectures defined as a directed graph where each edge has a capacity and receives a flow of information, as described by Waissi et al.~\cite{waissi1994network} (1994).
Precisely by non-negative and non-linear projection $\phi$ of flow computation capacity. 
The introduced operation aggregates the weights (i.e., attention maps) computed from the queries $(Q)$ and keys $(K)$ over the information values $(V)$, achieving a linear complexity with respect to the number of input tokens.
The authors define the capacity of conserved information flows as $\hat{I}\in \mathbb{R}^{n\times1}$, the incoming flow, and with $\hat{O}\in \mathbb{R}^{m\times1}$ the outgoing flow.
\revtwo{Therefore, the overall flow attention module $A_{Flow}$ can be formulated based on flow conservation principles~\cite{waissi1994network} (competition, aggregation and allocation) as reported in Equation~\ref{eq:att_flowformer}.}
\begin{equation}
    \label{eq:att_flowformer}
    \begin{split}
        & Competition: \hat{V} \in \mathbb{R}^{m\times d} = Softmax(\hat{O}) \odot V \\
        & Aggregation: A \in \mathbb{R}^{m\times d} = \frac{\phi(Q)}{I} \cdot \big ( \phi(K)^T \cdot \hat{V} \big ) \\
        & Allocation: A_{Flow} \in \mathbb{R}^{n\times d} = Sigmoid(\hat{I} \odot A) 
    \end{split}
\end{equation}

\revone{Furthermore, Bolya et al.~\cite{hydraattention} (2022), inspired by the use of decomposable kernel strategies~\cite{katharopoulos2020transformers} (linear in the number of tokens)  in order to reduce the computational cost of self-attention, proposed to take a step further by developing the Hydra Attention module. 
The proposed technique is based on the concept that increasing the number of attention heads in the MSA does not increase the computational cost. 
Consequently, aligning the number of attention heads with the features in a decomposable kernel strategy, the model achieves computational linearity in both tokens and features, resulting in a reduced computational cost of the attention layer.
Moreover, as reported in the original paper, this method is particularly effective for high-resolution images, a common bottleneck for high-demanding architectures.}

Huang et al.~\cite{huang2022orthogonal} (2022) propose a general-purpose backbone called Orthogonal Transformer (Ortho), which is able to reduce the computational cost of standard attention mechanism with an orthogonal self-attention ($A_{OSA}$). 
The attention is computed orthogonalizing the tokens within local regions, permuting them into token groups; the orthogonal matrix is defined as $O\in \mathbb{R}^{n\times n}$.
Finally, the MSA is computed group-wisely into the orthogonal space.
This solution leads to the possibility of computing the $A_{OSA}$ with a lower resolution with respect to the commonly used image space, reducing the overall computational complexity of self-attention of a factor $n_0$ equal to the number of groups into which the tokens are separated.
The authors also introduce a Positional MLP in order to incorporate
position information for arbitrary input resolutions into Ortho.
\revtwo{Therefore, the output of the attention module can formalized as reported in Equation~\ref{eq:att_ortho}; the authors employ the inverse orthogonal matrix after the MSA in order to recover the visual tokens from the orthogonal representations.}
\begin{equation}
    \label{eq:att_ortho}
    X_{out} = O^T \cdot X_{MSA}(O) 
\end{equation}

Recently, Liu et al.~\cite{liu2023ecoformer} (2023) propose to reduce the computational cost of self-attention by taking advantage of an extreme quantization technique.
\revtwo{Consequently, we characterize this study as CA + Q since, similar to previous models, as it aims to reduce the computing cost of the self-attention module by leveraging the use of a quantization (Q) methodology.}
The authors introduce EcoFormer, a binarized self-attention module that extensively reduces the multiply and accumulates (MACs) operations in standard attention layers in order to save a considerable on-chip energy footprint. 
More in detail, by defining with $D_p$ the number of dimensions for each attention head and $b$ the number of bits, the proposed solution, based on kernelized hash functions $H$, with $H\in \mathbb{R}^{D_p} \mapsto \{1, -1\}^b$, is particularly effective for edge devices, where the computational resources are a bottleneck for high capacity deep learning architectures.
\revtwo{Therefore, inspired by the idea of kernel-based linear attention~\cite{rahimi2007random}, the attention equation ($A_{Eco}$) can be formalized as follows.}
\begin{equation}
    \label{eq:att_ecof}
    A_{Eco} =  Softmax \bigg ( \frac{H(Q)^T \cdot H(K)}{\sqrt{d_k}} \bigg ) \cdot V
\end{equation}
The authors take advantage of an \textit{extreme} quantization scheme ($b=16$ hashing bit), which is able to represent feature vectors\footnote{Weights and operations in Deep Learning models are usually computed at floating point precision data type.} in binary codes. 
However, although the solution is demonstrated to be energy and memory saving, the binary compression requires specialized GPU kernels to be piratically deployed on edge devices; therefore, these limitations result in a bottleneck on the effective efficiency of EcoFormer self-attention modules when employed in real-case scenarios.

Finally, You et al.~\cite{you2023castling} (2023) introduced a novel framework named Castling-ViT, which aims to be a ViT structure composed of only linear terms.
To tackle this problem, the authors introduce a linear-angular attention module, where angular kernels are decomposed in linear terms, and the remaining high-order residuals are approximated with a depth-wise convolution and an auxiliary masked-self-attention operation, i.e., an attention module which only focuses on a limited number of patches.
Although the latter element still has a quadratic computing cost, the authors noted that it tends to converge to zero during the training phase, thus being worthless during the inference phase.
\revtwo{Moreover, by defining the angular kernel ($Sim(Q, K)$) as a similarity measurement function computed between the queries $Q$ and keys $K$, the linear-angular attention module ($A_{Cast}$) can be formulated as reported in Equation~\ref{eq:att_casting}.}
\begin{equation}
    \label{eq:att_casting}
    A_{Cast} = Sim(Q, K) \cdot V    
\end{equation}

\revone{In contrast to previous studies focused on efficient backbones for general-purpose applications, Cai et al.~\cite{cai2023efficientvit} (2023) propose EfficientViT, an efficient ViT architecture with linear computational cost designed to handle high-resolution dense prediction tasks such as semantic segmentation and super-resolution. 
The authors leverage the use of the ReLU-based global attention~\cite{katharopoulos2020transformers} to achieve both the global receptive field of ViT and linear computational complexity.}

\subsection{Pruning}
\label{sec:pruning_decomp}
This section reviews the state-of-the-art ViT pruning method based on the background notions and mathematical formulations introduced in Section \ref{sec:pruning_background}.
Similar to previous analysis, we review the proposed solutions to prune ViT architectures according to their release date. 
In particular, we will emphasize how pruning algorithms evolved in recent years over both strategies to compute the importance score and adaptive solutions for the pruning/preserving ratio in order to maximize the performance of the algorithm.

A preliminary study has been performed by Zhu et al.~\cite{zhu2021vision} (2021), who propose VTP, a vision transformer pruning method that is able to thin out ViT architectures while encouraging dimension-wise sparsity. 
The solution mainly focuses on MSA and FNN transformer structures in order to identify less informative features, i.e., by reducing the number of embedding/neuron dimensions via control coefficients.
As a common pruning strategy, VTP lays the groundwork on a feature importance score ($p$).
This strategy is based on learning at training time the soft pruned features $\hat{p}_{0/1} \in \{0, 1\}^\mathbb{R}$, while defining the hard pruned features $p_{0/1}^*\in \{0, 1\}^\mathbb{N}$ at inference time, based on a threshold value $\tau\in \{0, 1\}^\mathbb{R}$, i.e., $p_{0/1}^* = \hat{p}_{0/1} \geq \tau$.
\revtwo{Then, the hard pruned self-attention transformer block ($A^*_{VTP})$ can be formulated as reported in Equation~\ref{eq:pruning_vtp}, where $P_{0/1}$ is a diagonal matrix whose diagonal line is composed of $p_{0/1}^*$-elements, i.e., $P_{0/1}=diag(p_{0/1}^*)$.}
\begin{equation}
    \label{eq:pruning_vtp}
    A^*_{VTP}(Q^*, K^*, V^*) = P_{0/1} \cdot A(Q, K, V)
\end{equation}
Precisely, if the algorithm attributes a value of $p=0$ to the feature, it will be discarded; otherwise ($p=1$), it will be maintained.

Moreover, Tang et al.~\cite{tang2022patch} (2022), inspired by a previous approach on CNN architectures proposed in~\cite{liu2017learning}, introduce PS-ViT, a patch slimming framework to improve the efficiency of vision transformers structures. 
The proposed solution is motivated by the common problem of overparameterization and redundant information of deep learning algorithms.
Therefore, PS-ViT mainly focuses on identifying transformer patches with redundant information in order to discard them and accelerate the inference process.
As common pruning algorithms, in PS-ViT, each patch in the attention bock receives an importance score $p=\{0,1\}^\mathbb{N}$.
Differently from VTP, the $p_{0/1}$-values are learned during the backpropagation phase with a top-down procedure, i.e., from 
from the output layer to the input one.
Consequently, the pruned attention module $A^*_{PS-ViT}$ can be reformulated as reported in Equation~\ref{eq:pruning_psvit}, where $P_{0/1}$ is a diagonal matrix whose diagonal line is composed of $p_{0/1}$-elements.
\begin{equation}
    \label{eq:pruning_psvit}
    A^*_{PS-ViT}(Q^*,K^*,V^*) = P_{0/1}\cdot A(Q,K,V) 
\end{equation}
Precisely, if the algorithm attributes a value of $p=0$ to the patch, it will be identified as redundant and will be discarded; otherwise ($p=1$), it will be maintained. 
Moreover, Tang et al. propose a dynamic variant of the patch slimming algorithm, named DPS-ViT, which adaptively selects the non-redundant features at inference time depending on the input samples. 

Furthermore, Yu et al.~\cite{Yu2022WidthD} (2022), motivated by the fact that previous pruning strategies only focus on architecture width, develop the Width \& Depth Pruning (WDPruning) framework. 
The proposed solution reduces the architecture width via a learnable saliency score threshold, similar to previous works, while limiting the structure depth by introducing multiple classifiers inside the model.
WDPruning framework has the final objective of determining an optimal trade-off between accuracy and efficiency, with a shallower model from both depth and width perspectives.
Precisely, the width pruning is commonly based on a diagonal binary matrix $P_{0/1}$, while the threshold ($\tau$) dynamically updates the pruning ratio via the Augmented Lagrangian method at the training phase until reaching a predefined score.
Moreover, the depth pruning procedure has the objective of identifying the shallower classifier; this procedure is computed via several classifiers plugged into the architecture, which are evaluated at validation time in order to determine an optimal trade-off between estimation performances and efficiency.

Based on a different approach with respect to previous works, Yang et al.~\cite{yang2023global} (2023) propose a global\footnote{Differently from local pruning, the global one focuses on all the network's parameters in order to prune a fraction of them.} structural pruning criteria, which is able to guarantee parameter redistribution along the network.
The authors propose to compute the importance score ($p$) as the Hessian matrix norm of the loss, which, in contrast to local pruning strategies focusing on specific layers/neurons, leads to a global pruning of the overall architecture.
Moreover, this solution, applied to the DeiT-Base model, enables the generation of new efficient ViT models named NViT.

\revone{Recently, Rao et al.~\cite{rao2023dynamic} (2023), extend the previously developed DynamicViT framework \cite{rao2021dynamicvit} (2021), utilized to increase sparsity and accelerate the inference of general neural network structures over ViT architectures.}
The framework bases the groundwork on~\cite{tang2022patch}, where only a subset of image patches are needed for the final prediction. 
Therefore, in DynamicViT, the authors propose to progressively and dynamically\footnote{\revtwo{In a dynamic pruning framework, during the training phase, the model learns how to select the most/least informative tokens (based on a score) so that at the inference phase, the model is able to classify them according to the specific input features, thus reducing the computation and increasing the model’s throughput. In contrast, the pruning percentage of the reference model is usually chosen a priori.}} prune less informative tokens hierarchically based on a prediction module. 
The latter structure is placed between Transformer blocks in order to individuate and discard less informative tokens, while a threshold value is used as a ratio to determine the percentage of the information to be preserved.
Moreover, in order to minimize the influence on performance drop caused by spatial sparsification, the authors also leverage knowledge distillation techniques in order to guide the pruned-student model to the teacher behavior.
Precisely, the original backbone network is used as the teacher model, while an equal structure is dynamically pruned in order to obtain the student variant. 
\revone{In addition, motivated by DynamicViT, researchers investigate methods for handling adaptive inference approaches in ViT structures based on pruning/merging of less informative tokens; we provide an in-depth study of such methodologies, such as~\cite{yin2022vit, kong2022spvit, bolya2023tome}, in the supplementary material Section~\ref{subsec:dynamic_inference}.}

Finally, Yu and Xiang~\cite{yu2023x} (2023) proposed X-Pruner, a layer-wise pruning algorithm that leverages eXplainable AI (XAI) principles in the pruning strategy.
The baseline idea is to identify and prune less contributing attention units from an explainability perspective.
Specifically, the authors introduce a learnable explainability-aware mask ($M$) that can be used to prune or unprune the model based on an adaptive threshold value ($\theta$).
\revtwo{Moreover, by defining the threshold ratio $r$, two hyperparameter values respectively set to $h_1=10$ and $h_2=500$, and with $\Phi$ a function that returns the top $(1 - r)\%$ sorted elements of $M$, the desired mask $\hat{M}$ for a generic layer can be formally defined as reported in Equation~\ref{eq:xpruner_mask}.}
\begin{equation}
    \label{eq:xpruner_mask}
    \hat{M} = \begin{cases}
    M tanh(h_1 (M-\theta)) & \textrm{if $M \in \Phi(M|1-r)$}  \\ 
    h_2 tanh(h_1 (M-\theta)) & \textrm{\ otherwise}
    \end{cases}
\end{equation}

\subsection{Knowledge Distillation}
\label{sec:knowledge_distill}
This section reviews knowledge distillation learning techniques specifically designed to prove and design lightweight ViT models. 
Please refer to Section \ref{sec:KD_background} for background information.
Similar to previous analysis, we review KD solutions for ViT architectures according to their release date. 
In particular, we will emphasize how the well-known vanilla KD strategies have been first adapted to ViT architectures and successively improved in order to maximize the student's ability to mimic the teacher's behavior. 

Touvron et al. \cite{touvron2021training} (2021) have been the first to explore the KD learning strategy for ViT architectures by introducing a token-based strategy denoted by DeiT.
The proposed transformer-specific approach is based on distillation through attention, i.e., the authors add a distillation token into the self-attention module, which aims to reproduce the class (hard) label estimated by the teacher. 
Therefore, the token will interact with both student attention and layers and teacher labels learning the hard labels during the back-propagation.

Moreover, motivated by the high computational cost of ViT and their difficult application on edge and low-power devices, Hao et al.~\cite{hao2022learning} (2022) explore the patch-level information present in transformers modules in order to propose a fine-grained manifold distillation method. 
This manifold strategy learns a smooth manifold ($\mathcal{M}$) embedded in the original feature space to construct low-dimensional features ($\psi(F)$).
Precisely, the authors train the student model to match a pretrained teacher in the patch-level manifold space.
The proposed solution led to the introduction of the manifold distillation loss function ($\mathcal{L}_{mf}$), reported in Equation~\ref{eq:kd_loss_manifold}. 
\begin{equation}
    \label{eq:kd_loss_manifold} 
    \mathcal{L}_{mf} = ||\mathcal{M}(\psi(F_S)) - \mathcal{M}(\psi(F_T))||^2_F
\end{equation}

Wu et al.~\cite{wu2022tinyvit} (2022) propose a new family of architectures named TinyViT. 
The structure of these models is obtained via a constrained local search~\cite{hoos2004stochastic} algorithm in the model space spanned by multiple constrained factors and is trained with a memory-efficient KD strategy.
Precisely, the latter procedure focuses on storing sparse\footnote{Selecting only the top-K soft values from the classification vector.} soft-labels of deep and heavy pretrained models into storage devices.
Consequently, at training time, it will be possible to reuse the stored information in order to replicate the vanilla KD procedure while omitting the forward computation and memory occupation of the large teacher model.
Moreover, the search algorithm used to generate the TinyViT architecture's family is constrained to computationally demanding ViT elements such as the depth of the model and patch size.

Differently to previous methods that apply the KD strategy from a ViT teacher to a ViT student model, Chen et al.~\cite{chen2022dearkd} (2022), motivated by the high number of real samples needed to train ViT architectures, propose a two-stage learning framework, named Data-efficient EARly Knowledge Distillation (DearKD).
The first stage is composed of a vanilla KD learning strategy where, inspired by~\cite{Cordonnier2020On}, CNN features extracted from both the inner and classification layers of the models are distilled with transformer tokens. 
Moreover, in the second stage, the ViT student model is trained without distillation.
However, in the case of a limited number (or data-free) of real samples, during this second phase, the authors introduce a boundary-preserving intra-divergence loss based on DeepInversion~\cite{yin2020dreaming}, which helps the learning procedure to keep the easiest positive samples away from others in the latent space without changing its boundaries.

Similar to Chen et al., Ren et al.~\cite{ren2022co} (2022), propose
a KD strategy that is not based on distilling the knowledge from teacher and student models with similar architecture structures, i.e., two ViT, but via different structures.
Precisely, Ren et al. rely on the idea that teacher models with different inductive biases could extract different features, i.e., looking at the samples from different perspectives and consequently co-advising the student transformer in order to improve its estimation performances.
Therefore, the proposed strategy is based on vanilla KD procedure applied to two lightweight teachers, a CNN and an involution neural network (INN)\footnote{INN, introduced in~\cite{li2021involution},  has the ability to relate long-range spatial relationship in an image thanks to involution kernels which are shared across channels; differently to convolution kernels which are limited to channels.}, which focuses on different input samples' features, despite that they are trained on the same dataset, leading to an improvement of the ViT student (CivT) accuracy at training phase.

Zhang et al.~\cite{zhang2022minivit} (2022) propose MiniViT, a compression strategy based on KD to generate \textit{Mini}-versions of well-known ViT models, such as Mini-Swins and Mini-DeiTs respectively, from the original (teacher) Swin and DeiT transformers architectures, 
The proposed strategy lays the groundwork on the weight multiplexing process, which is applied to both attention matrices and feed-forward networks.
More in detail, the process focuses on weight sharing, transformation, and distillation from the teacher model  ($t$) to the student ($s$) one in order to improve training stability and model performance.
The described procedure is applied over both attention matrices and feed-forward networks.
However, due to the dimensionality reduction of the student model with respect to the teacher one, the authors apply cross-entropy ($CE$) losses on the relations among queries (Q), keys (K), and values (V) of the MSA by defining the self-attention distillation loss $\mathcal{L}_{att}$ and to transformer hidden states by defining the hidden-state distillation loss as $\mathcal{L}_{hddn}$ as respectively reported in Equation~\ref{eq:l_att_KD} and Equation~\ref{eq:lhidd_kd}.
Precisely, by denoting with $M_1$, $M_2$, and $M_3$ the Q, K, and V matrices with equal sizes ($d_q = d_k = d_v$), with $N$ the number of patches, and $X_{out}$ the output features of the feed-forward network as described in Section~\ref{sec:background_attention_attention}, it is possible to compute the two previous equations as following reported.
\begin{equation}
    R_{i,j\in \{1,2,3\}} = Softmax \bigg(\frac{M_i \cdot M_j^T}{\sqrt{d_k}} \bigg)
\end{equation}
\begin{equation}
    \label{eq:l_att_KD}
    \mathcal{L}_{att} = \frac{1}{9N}\cdot\sum_{p=1}^N\sum_{i,j\in \{1,2,3\}}CE(R_{ij,p}^s, R_{ij,p}^t)
\end{equation}
\begin{equation}
    \label{eq:lhidd_kd}
    \mathcal{L}_{hddn} = \frac{1}{N} \sum_{p=1}^N CE(R^s_{X_{out},p},R^t_{X_{out},p})
\end{equation}
The train process is finally computed by combining the counterbalance of the vanilla distillation loss function ($\mathcal{L}_{Dstl}$) and the two just introduced attention distillation losses.

On a different application scenario, Lin et al.~\cite{lin2023supervised} (2023) focus on ViT models applied for few-shot learning (FSL) tasks on small datasets. 
Under these settings, ViT tends to overfit and suffers from severe performance degradation due to the high number of trainable parameters.
Consequently, the authors focus on a KD learning strategy named supervised masked KD (SMKD), in which the student model only learns from a masked input sample, i.e., a reduced number of patches. 
Moreover, the authors introduce an intra-path loss function $\mathcal{L}_{patch}$ which compares the student ($s$) and teacher ($t$) embedding vectors by computing the cross-entropy loss of $s/t-$matching patches.
The introduced strategy is finally completed by adding the vanilla KD learning procedure to the overall learning procedure.

\subsection{Quantization}
\label{sec:quantization}
Based on the mathematical background introduced in Section \ref{sec:quantization_background}, and motivated by the fact that ViTs are both memory and computation expensive during inference, this section reviews multiple researches focused on ViT quantization strategies specifically designed to reduce the costs of memory and computation.
Similar to previous analysis, we review these efficient solutions for ViT architectures according to their release date.
In particular, we will focus on general and hardware-specific quantization strategies designed to closely match ViT full precision data distribution with a lower bit-width.

Liu et al.~\cite{liu2021post} (2021) have been the first to explore post-training quantization for ViT architecture. 
The proposed compression strategy, based on mixed-precision weights, is formulated as an optimization problem without taking into account any training or fine-tuning process.
Basically, the problem has the objective of finding the optimal low-bit quantization intervals for both weights and inputs in order to reduce both memory storage and computational costs. 
The authors focus on the FNN and MSA modules of ViT, aiming to assign the lowest possible bit-width to each attention module in order to maximize the prediction similarity between the full-precision model and the quantized one.

Similarly to Liu et al., Yuan et al.~\cite{yuan2022ptq4vit} (2022) propose an efficient framework for post-training quantization named PTQ4ViT.
The main advantage of the proposed solution, with respect to the previous one, is the use of a twin uniform quantization strategy, which separately quantifies the negative and positive values in two ranges: $R_1$ and $R_2$, respectively. 
This choice is mainly due to the values achieved after softmax layers and GeLU activation functions of the attention block since their data distribution is very unbalanced and consequently very difficult to quantify.
Therefore, based on Equation~\ref{eq:quant_range_eq_split}, and by defining the quantization intervals (scaling factors) $\Delta_{R_1}$ and $\Delta_{R_2}$ respectively for the two ranges $R_1$ and $R_2$, we can define  the \textit{k}-bit twin uniform quantization as reported in Eqution~\ref{eq:ptq4vit}.
\begin{equation}
    \label{eq:ptq4vit}
    T(x, \Delta_{R_1}, \Delta_{R_2}) =  \begin{cases}
    \Psi_{k-1}(x, \Delta_{R_1}) & \textrm{\ if $x\in R_1$} \\ 
    \Psi_{k-1}(x, \Delta_{R_2}) & \textrm{\ otherwise}
    \end{cases}
\end{equation}
Moreover, in order to guide the optimal scaling factors for each layer for ViT, the authors propose to use the Hessian-guided metric to determine the quantization parameters.

However, Ding et al.~\cite{ding2022towards} (2022), motivated by a notable accuracy drop at ultra-low bit-widths quantization, i.e., 4-bit, when Hessian-guided metric is employed to measure the quantization loss, propose a different approach named Accurate Post-training Quantization framework for Vision Transformer (APQ-ViT).
Precisely, the authors focus on the design of a Matthew-effect\footnote{In mixed-precision, the Matthew-effect~\cite{hong2022dropnas} is the situation in which layers with higher bit-widths would be trained maturely earlier, while the others with lower bit-widths may never have the chance to express the desired function.} Preserving Quantization (MPQ) for the softmax function of the attention block.
More in detail, APQ-ViT is composed of two phases: a quantization loss based on a unified Blockwise Bottom-elimination Calibration scheme to optimize the calibration metric and the MPQ specifically tailored for ViT models.
\revtwo{Precisely, by defining the softmax function as $softmax(\cdot)$, the scaling factor ($\Delta_{MPQ}$) can be defined as reported in Equation~\ref{eq:apq-vit-SF} and the MPQ function as reported in Equation~\ref{eq:apq-vit-mpq}.}
\begin{equation}
    \label{eq:apq-vit-SF}
    \Delta_{MPQ} = \frac{max(softmax(\cdot))}{2^k - 1}
\end{equation}
\begin{equation}
    \label{eq:apq-vit-mpq}
    \Psi_k(\cdot, \cdot) = Clamp \bigg(Round \bigg(\frac{softmax(\cdot)}{\Delta_{MPQ}} \bigg), 0, 2^k -1 \bigg)
\end{equation}

Moreover, Liu et al.~\cite{liu2023noisyquant} (2023) propose a plug-and-play quantizer-agnostic enhancement method for post-training ViT activation functions quantization.
The proposed strategy, namely NoisyQuant, aims to improve previous quantization methods by adding a fixed NoisyBias ($Nb$) sampled from the Uniform distribution.
Precisely, given the output distribution of the GeLU activation function ($X$), the NoisyQuant strategy can be obtained as $X+Nb$.
This operation, computed before the quantization, flattens the data peaks, making the overall compression process more friendly.
Moreover, the authors demonstrate that this sort of \textit{soft-bounds} applied to the data distribution leads to a quantized output that closely follows the original data distribution with a lower bit rate.

Quantization methods are not only focused on compressing specific attention layers and respective activation functions, i.e., general ViT modules, in order to mimic original data distribution with lower precision data types; some works also focus on ViT compression strategies for specific hardware devices.
This choice is mainly due to the real-world application of quantized models; this fact is particularly true in some scenarios like the binary compression proposed in Liu et al. \cite{liu2023ecoformer} (EcoFormer).
Precisely, when analyzing the limitations of the proposed method, the authors state that although EcoFormer is more efficient than previous solutions thanks to the binarization, in real-world applications, such as when inferring on GPU platforms, specific GPU kernel implementation, i.e., CUDA\footnote{CUDA is a software platform that enables accelerated GPU computing on multiple operating systems.} operations would be required to take advantage of the proposed solution. 
This fact is due to the GPU device, which is unable to perform binary computations without allocating (in any case) a floating-point operation.

In this settings, Lit et al.~\cite{lit2022auto} (2022) propose a framework, named Auto-ViT-Acc, which is specifically designed for quantizing ViT architectures to infer on FPGA\footnote{An FPGA is a programmable logic device composed by an integrated circuit whose logical processing functionalities are programmable.} powered devices.
The framework, which takes advantage of the quantization function introduced in \cite{chang2021mix}, is applied only on FNN module of the attention block in order to increase the FPGA resource utilization and speed the inference process.

Moreover, differently from all the previous approaches, Yu et al.~\cite{yu2023boost} (2023) propose GPUSQ-ViT, a GPU-friendly framework that incorporates multiple compression techniques. 
The proposed mixed strategy leverages the use of the knowledge distillation learning technique during a preliminary pruning phase and the subsequent aware-training quantization.
Precisely, in a KD learning strategy, the full precision model, also used as a teacher, is first pruned with 16-bit floating point weights and then quantized to a fixed-point (precision) in order to obtain the final student model.

\section{Efficient Vision Transformer Performances}
\label{sec:results_and_applications}
In this section, we review and compare the estimation performances of previously described efficient ViT strategies. 
More in detail, following the four selected efficient categories (CA, P, KD, and Q), we compare all the models on the ImageNet~\cite{imagenet} classification task.
\revtwo{Precisely, all the reported values are extracted from the original papers and refer to the architectures trained end-to-end on the ImageNet1K dataset.}
The ImageNet1K dataset is composed of $1.3$M of training and $50$K validation images covering common $1$K classes at a resolution of $224\times 224$ pixels.
Moreover, for the CA efficient strategy, we report into the supplementary material, Section~\ref{subsec:coco_adek}, the obtained results also on two other datasets, i.e., the COCO~\cite{lin2014microsoft} object detection and instance segmentation dataset, and the ADE20K~\cite{zhou2017scene} semantic segmentation datasets.

We evaluate the compared models with respect to the accuracy metric (Top-1) for the classification, average precision (AP), i.e., AP$^{box}$ and AP$^{mask}$ for the object detection and instance segmentation, and mean intersection over union (mIoU) for the semantic segmentation.
\revone{Moreover, we also report into Tables~\ref{tab:results_on_imagenet1k_CA}~\ref{tab:results_on_imagenet1k_P}~\ref{tab:results_on_imagenet1k_KD}\footnote{\revone{We report DA techniques for  CA, P, and KD categories since Q methodologies are usually compared only on three pretrained models.}} the data augmentation (DA) techniques used to train the compared models. 
More in detail, we identify as baseline DA strategy the one proposed in~\cite{touvron2021training} (indicated as \cmark), which includes Rand-Augment, Mixup, CutMix, and random erasing operations, excluding Repeated Augmentation and Stochastic Depth.
In the tables, symbols \cmark\cmark~and \xmark~ indicate cases with more or less DA techniques compared to the baseline.}
Furthermore, we report the number of trainable parameters (\#Par.) and floating point operations (FLOPs) as efficiency metrics for AC, P, and KD strategies, as well as the weight/activation bit-widths (\#Bit) and model size (Size) for Q methods, in order to quantify and evaluate the impact of the executed optimizations.
Moreover, inspired by Li et al.~\cite{li2022rethinking}, we introduce a metric that is able to measure how much a model is efficient with respect to a reference baseline; we define the Efficient Error Rate (EER) as reported in Equation~\ref{eq:eer_metric}.
\begin{equation}
    \label{eq:eer_metric}
    EER = \frac{1}{||i||} \cdot \sum_{i} \bigg( \frac{M_i}{R_i} \bigg)
\end{equation}
Where $i\in\{\#Par., FLOPs, ...\}$ is the set of efficiency metrics, $M_i$ and $R_i$ are respectively the $i-$values for the efficient model under analysis and for the reference one.  
Consequently, the more efficient the proposed model, with respect to the baseline model chosen for the specific task, the lower the EER value will be.
\revone{Furthermore, in the supplementary material (Section~\ref{subsec:ERR_metric}), we present a more in-depth examination of the proposed EER metric in order to highlight its applicability and flexibility across diverse real-world tasks.}
Moreover, the EER metric is solely measured in the classification task since it is an application scenario in which all the identified efficient categories (CA, P, KD, and Q) are compared.
More in detail, in the following tables, we report as percentage value the EER computed by choosing as baseline ($R$) the ViT-B/16.~\cite{dosovitskiy2020image} architecture.
This choice has been derived from the fact that the reference architecture is the shallower non-optimized model among the ViT milestones originally proposed by Dosovitskiy et al. and thus best approximates the efficient solutions under consideration in terms of \#Pram., FLOPs, \#Bit, and Size. 
Generally speaking, the ViT-B/16 model could be considered as the upper bound of the efficient/lightweight architectures distribution in analysis.
Such architecture counts $86.6$ million (M) of trainable parameters, $17.6$ gigaflops (G) of operations at 32-bit precision, and a model size of $330$ megabyte (MB).
We underline that in order to have a fair and comprehensive understanding of previously analyzed methods with respect to the chosen evaluation metrics, models that have not been trained/tested according to the previously reported criteria, i.e., with different image resolutions or evaluation datasets, will not be present in the following tables; at the same time, all the values that are not given in the respective papers will be denoted with the $-$ symbol.

The rest of the section is organized as follows: Section~\ref{subsec:results_of_CA}, Section~\ref{subsec:results_of_P}, Section~\ref{subsec:results_of_KD}, and Section~\ref{subsec:results_of_Q} respectively reviews the estimation performances of CA, P, KD, and Q strategies.
Finally, Section~\ref{subsec:results_of_CAPKDQ_calass} compares all the strategies that best perform on the well-known classification scenario, i.e., over the $1$K classes of the ImageNet benchmark dataset.  

\subsection{Results of Compact Architectures strategies}
\label{subsec:results_of_CA}
In this section, we compare the classification performances of compact architectures (CA) previously introduced in Section~\ref{sec:compact_arch}.
Moreover, we report in the supplementary material Section~\ref{subsec:coco_adek} their estimation performances over other object detection, instance, and semantic segmentation tasks.
We compare the classification performances of CA methods over the ImageNet1K dataset in Table~\ref{tab:results_on_imagenet1k_CA}\footnote{Please refer to the supplementary material Section~\ref{subsec:graphs} for a graphical representation of the model's distributions regard to accuracy-EER values.}.
\begin{table}[t]
    \centering
    \footnotesize
    \caption{Quantitative comparison of \textbf{CA} models on ImageNet1K classification dataset. The best results are in bold, and the best (trade-off) efficient model is highlighted in gray.}    
    \begin{tabular}{ l | c c c c | c }
         \multirow{2}*{Model} & \#Par. & FLOPs & Top-1 & EER & \multirow{2}*{DA} \\
          & [M] & [G] & [\%] & [\%]\\
         \hline
         PVT-Tiny & 13.2 & 1.9 & 75.1 & 13.0 & \multirow{4}*{\revone{\cmark}} \\
         PVT-Small & 24.5 & 3.8 & 79.8 & 24.9 &  \\
         PVT-Medium & 44.2 & 6.7 & 81.2 & 44.5 &  \\
         PVT-Large & 61.4 & 9.8 & 81.7 & 63.3 & \\
         \hline
         Swin-T & 29 & 4.5 & 81.3 & 29.5 & \multirow{3}*{\revone{\cmark\cmark}}\\
         Swin-S & 50 & 8.7 & 83.0 & 53.5 & \\
         Swin-B & 88 & 15.4 & 83.5 & 94.5 & \\
         \hline
         SOFT-Tiny & 13 & 1.9 & 79.3 & 12.9 & \multirow{5}*{\revone{\xmark}}\\
         SOFT-Small & 24 & 3.3 & 82.2 & 23.2 & \\
         SOFT-Medium & 45 & 7.2 & 82.9 & 46.4 &  \\
         SOFT-Large & 64 & 11.0 & 83.1 & 68.2 & \\
         SOFT-Huge & 87 & 16.3 & 83.3 & 96.5 &  \\
         \hline
         PoolFormer-S12 & 12 & 1.9 & 77.2 & 12.3 & \multirow{5}*{\revone{\cmark}}\\
         PoolFormer-S24 & 21 & 3.5 & 80.3 & 22.1 & \\
         PoolFormer-S36 & 31 & 5.1 & 81.4 & 32.4 & \\
         PoolFormer-M36 & 56 & 9.0 & 82.1 & 57.9 & \\
         PoolFormer-M48 & 73 & 11.8 & 82.5 & 75.7 & \\
         \hline
         PVTv2-B2-LiSRA & 22.6 & 3.9 & 82.1 & 24.0 & \revone{\cmark} \\
         \hline
         MViTv2-T & 24 & 4.7 & 82.3 & 27.2 & \multirow{4}*{\revone{\cmark}} \\
         MViTv2-S & 35 & 7.0 & 83.6 & 40.1 & \\
         MViTv2-B & 52 & 10.2 & 84.4 & 59.0 & \\
         MViTv2-L & 218 & 42.1 & \textbf{85.3} & 245.5 & \\
         \hline
         XCiT-T12/8 (SimA) & 7 & 4.8 & 79.4 & 17.7 & \multirow{2}*{\revone{\cmark}}\\
         XCiT-T12/16 (SimA) & 26 & 4.8 & 82.1 & 28.6 & \\
         \hline
         Ortho-T & \textbf{3.9} & 0.7 & 74.0 & \textbf{4.2} & \multirow{4}*{\revone{\cmark}}\\
         Ortho-S & 24.0 & 4.5 & 83.4 & 26.6 & \\
         Ortho-B & 50.0 & 8.6 & 84.0 & 53.3 & \\
         Ortho-L & 88.0 & 15.4 & 84.2 & 94.5 & \\
         \hline
         Flowformer & - & 6.3 & 80.6 & - & \revone{-} \\
         \hline
         Castling-LeViT-128 & 10.5 & \textbf{0.49} & 79.6 & 7.4 & \multirow{8}*{\revone{-}} \\
         Castling-LeViT-192 & 12.7 & 0.82 & 81.3 & 9.6 & \\
         Castling-LeViT-256 & 22.0 & 1.4 & 82.6 & 16.7 & \\
         Castling-LeViT-384 & 45.8 & 2.9 & 83.7 & 34.7 & \\
         \cellcolor{gray!30}{Castling-MViTv2-T} & \cellcolor{gray!30}{24.1} & \cellcolor{gray!30}{4.5} & \cellcolor{gray!30}{84.1} & \cellcolor{gray!30}{26.7} & \\
         Castling-MViTv2-S & 34.7 & 6.9 & 84.6 & 39.6 & \\
         Castling-DeiT-B & 87.2 & 17.3 & 84.2 & 99.5 & \\
         Castling-MViTv2-B & 51.9 & 9.8 & 85.0 & 57.8 & \\
         \hline
         \revone{EfficientViT-B1} & \revone{9.1} & \revone{0.52} & \revone{79.4} & \revone{6.7} & \multirow{3}*{\revone{\cmark}} \\
         \revone{EfficientViT-B3} & \revone{49} & \revone{4.0} & \revone{83.5} & \revone{39.6} &  \\
         \revone{EfficientViT-L1} & \revone{53} & \revone{5.3} & \revone{84.5} & \revone{45.6} &  \\
         \hline
    \end{tabular}
    \label{tab:results_on_imagenet1k_CA} 
\end{table}
Based on the reported accuracy values, we can notice that the MViTv2-L model has obtained the highest estimation performances, equal to $85.3\%$.
In this survey, however, we are interested in efficient DL techniques, i.e., in solutions that aim to find the best trade-off between the model's accuracy and the introduced EER metric.
Moreover, MViTv2-L is the model with the highest number of trainable parameters, FLOPs, and, consequently, the EER value.
Therefore, the MViTv2-L model might be considered an outlier in comparison to all other reported methods when looking for the Paretian optimality\footnote{Paretian optimality/efficiency is an economic theory that describes a situation in which an improvement in the scenario of one variable cannot be achieved without negatively affecting another variable. 
Therefore, given our set of compared models, the Paretian efficiency is used to identify the model in which a higher accuracy cannot be reached without worsening the EER value, i.e., by establishing the best accuracy-EER trade-off.} between accuracy and EER.
Moreover, based on the same architecture, the Castling-MViTv2-B with an EER of $4,2\times$ lower and similar estimation performances ($85.0\%$) should be preferred.

Differently, in the application scenario in which hardware-constrained devices such as embedded single board PC are required, the Ortho-T model with only $3.9$M of trainable parameters and also a restricted amount of FLOPs operations ($0.7$G) should be considered. 
\revone{Similarly, we can notice that also XCiT-T12/8 with SimA optimization, EfficientViT-B1, and Castling-LeViT-128 could be viable solutions for an hardware-constrained scenario.
In particular, the two latter models, which only counts $0.52$G$/0.49$G FLOPs, an EER of $6.7\%/7.4\%$, and estimation performances comparable to heavier models (Top-1 equal to $79.4\%/79.6\%$), could be valuable architectures in scenarios where real-time inference performances are needed.}

Moreover, by comparing the two variants of PVT architectures, the contribution provided by the pooling operation, which is widely employed in many CA methods such as PoolFormer, PVTv2, and MViT, can be highlighted.
More in detail, PVTv2, based on a linearized version of the SRA self-attention, originally introduced in PVT, is capable of achieving equivalent estimation performances of PVT-Medium with a comparable EER value of PVT-Sall. 

\revone{Finally, for the CA efficient category, we identify the Paretian optimality between accuracy and the EER evaluation metric with the Castling-MViTv2-T model (even if no DA techniques are provided into the original study).}
Such architecture generates accurate estimations, i.e., a Top-1 accuracy of $84.1\%$, which is almost comparable ($-1.2\%$) to the $85.3\%$ of the heavier MViTv2-L but with an EER gain of $+9.2\times$.
Moreover, when compared with models with similar EER values, such as Swin-T ($29.5\%$), MViTv2-T ($27.2\%$), and Ortho-S ($26.6\%$), the identified architecture with an EER of $26.7\%$, obtains more accurate performances, with an average accuracy boost of $+1.8\%$.

\subsection{Results of Pruning strategies}
\label{subsec:results_of_P}
In this section, we compare the estimation performances of the pruning strategies, which are mainly introduced in Section~\ref{sec:pruning_decomp} and formalized in Section~\ref{sec:pruning_background}.
We remind that P strategies aim to minimize the number of active connections and neurons in a neural network by setting their weight to zero. 
Usually computed due to an overparametrization of DL architectures at training time, this process will not affect the number of model parameters (\#Par.) since they will still be saved in the network graph even with a zeroed weight. 
In contrast, this strategy will be beneficial for the inference time by lowering the amount of (non-zeroed) multiplications required to generate the final prediction.
Consequently, in the results obtained by the compared strategies, which are reported in Table~\ref{tab:results_on_imagenet1k_P}, it is possible to notice that P methodologies are typically applied to well-known architectures such as Swin and DeiT, where the same number of parameters correlates to a lesser amount of FLOPs (please refer to Table~\ref{tab:results_on_imagenet1k_CA} and Table~\ref{tab:results_on_imagenet1k_KD} for their values).
More in detail, taking the X-Pruner strategy as an example, it is possible to achieve an average reduction of  $29.9\%$ and $51.3\%$ respectively on Swin and DeiT models with limited average accuracy (Top-1) reduction of $-0.8\%$ and $-2.6\%$.

Based on the reported results, it can be noticed that more accurate strategies (Top-1 $\geq 83.5\%$) such as DP/DPS and DynamicViT take advantage of the LV-ViT~\cite{jiang2021all} structure, which is, as reported in the original paper and even without pruning, able to achieve better and faster estimation than Swin and DeiT architectures.
Furthermore, this fact can also be easily noticed when comparing PS and DPS over DeiT and LV-ViT at the same dimensionality class, i.e., S/S and B/M.
However, in order to determine the Paretian optimality between accuracy and EER, we identify the DynamicViT pruning technique applied to the LV-ViT-S/07 architecture as the best trade-off among the analyzed efficient solutions.
\revone{Moreover, as shown in Table~\ref{tab:results_on_imagenet1k_P}, the  DynamicViT-LV-S/07, which leverages a baseline (\cmark) DA strategy, is able to obtain high estimation performances (Top-1 $=83.0\%$) while maintaining a low EER ($28.9\%$) when compared to all the other reported methods.}
In contrast, from a purely efficient perspective, the X-Pruner strategy applied to the DeiT-Ti structure provides discrete performances with a very low EER, making it a suitable option in situations that require extremely restricted computational requirements.

\begin{table}[t]
    \centering
    \footnotesize
    \caption{Quantitative comparison of \textbf{P} models on ImageNet1K classification dataset. The best results are in bold, and the best (trade-off) efficient model is highlighted in gray. FLOPs$^*$ values represent the amount of non-zeroed operations.}
    \begin{tabular}{l | c c c c | c }
         \multirow{2}*{Model} & \#Par. & FLOPs$^*$ & Top-1 & EER & \multirow{2}*{DA}\\
         & [M] & [G] & [\%] & [\%] \\
         \hline
         VTP-DeiT-B ($\tau=0.2$) & 86.4 & 13.8 & 81.3 & 88.5 & \multirow{2}*{\revone{\cmark}} \\
         VTP-DeiT-B ($\tau=0.4$) & 86.4 & 10.0 & 80.7 & 77.9 \\
         \hline
         PS-DeiT-S & 22 & 2.6 & 79.4 & 20.1 & \multirow{8}*{\revone{\cmark\cmark}}\\
         DPS-DeiT-S & 22 & 2.4 & 79.5 & 19.5 & \\
         PS-DeiT-B & 87 & 9.8 & 81.5 & 77.3 & \\
         DPS-DeiT-B & 87 & 9.4 & 81.6 & 76.2 & \\
         PS-LV-ViT-S & 26 & 4.7 & 82.4 & 32.8 & \\
         DPS-LV-ViT-S & 26 & 4.5 & 82.9 & 32.3 & \\
         PS-LV-ViT-M & 56 & 8.6 & 83.5 & 56.7 & \\
         DPS-LV-ViT-M & 56 & 8.3 & 83.7 & 55.9 & \\
         \hline
         NViT-T & 6.9 & 1.3 & 76.2 & 7.6 & \multirow{4}*{\revone{\cmark}}\\
         NViT-S & 21 & 4.2 & 82.2 & 24.1 & \\
         NViT-H & 30 & 6.2 & 82.9 & 34.9 & \\
         NViT-B & 34 & 6.8 & 83.3 & 38.9 & \\
         \hline
         X-Pruner-DeiT-Ti & \textbf{5} & \textbf{0.6} & 71.1 & \textbf{4.6} & \multirow{5}*{\revone{\cmark\cmark}}\\
         X-Pruner-DeiT-S & 22 & 2.4 & 78.9 & 19.5 & \\
         X-Pruner-DeiT-B & 87 & 8.5 & 81.2 & 74.1 & \\
         X-Pruner-Swin-T & 29 & 3.2 & 80.7 & 25.8 & \\
         X-Pruner-Swin-S & 50 & 6.0 & 82.0 & 45.9 & \\
         \hline
         DynamicViT-LV-S/05 & 26.9 & 3.7 & 82.0 & 26.0 & \multirow{4}*{\revone{\cmark}}\\
         \cellcolor{gray!30}{DynamicViT-LV-S/07} & \cellcolor{gray!30}{26.9} & \cellcolor{gray!30}{4.6} & \cellcolor{gray!30}{83.0} & \cellcolor{gray!30}{28.9} & \\
         DynamicViT-LV-M/07 & 57.1 & 8.5 & 83.8 & 57.1 & \\
         DynamicViT-LV-M/08 & 57.1 & 9.6 & \textbf{83.9} & 60.2 & \\
         \hline
    \end{tabular}
    \label{tab:results_on_imagenet1k_P} 
\end{table}

\subsection{Results of Knowledge Distillation strategies}
\label{subsec:results_of_KD}
This section compares the estimation performances of student models trained under knowledge distillation learning techniques.
The following analysis is mainly concerned with the understanding of how novel or existing architecture (student) can benefit from the knowledge of an external supervisor  (teacher) in order to generate more accurate estimations without significantly increasing the \#Par. and FLOPs.
The findings obtained by compared models, which are mainly introduced in Section~\ref{sec:knowledge_distill} and formalized in Section~\ref{sec:KD_background}, are quantitatively reported in Table~\ref{tab:results_on_imagenet1k_KD}.

\begin{table}[t]
    \centering
    \footnotesize
    \caption{Quantitative comparison of student models trained with reviewed \textbf{KD} strategies on ImageNet1K classification dataset. The best results are in bold, and the best (trade-off) efficient model is highlighted in gray.
    }    
    \begin{tabular}{ l | c c c c | c }
         \multirow{2}*{Model (student)} & \#Par. & FLOPs & Top-1 & EER &  \multirow{2}*{DA} \\
          & [M] & [G] & [\%] & [\%] \\
         \hline
         DeiT-Ti & 5 & \textbf{1.3} & 74.5 & 6.6 & \multirow{3}*{\revone{\cmark\cmark}} \\
         DeiT-S & 22 & 4.6 & 81.2 & 25.7 & \\
         DeiT-B & 87 & 17.6 & 83.4 & 100.2 & \\
         \hline
         Manifold-DeiT-Ti & 5 & \textbf{1.3} & 76.5 & 6.6 & \multirow{3}*{\revone{\cmark}}\\
         Manifold-DeiT-S & 22 & 4.6 & 82.2 & 25.7 & \\
         Manifold-Swin-T & 29 & 4.5 & 82.2 & 29.5 & \\
         \hline
         TinyViT-5M & 5.4 & \textbf{1.3} & 79.1 & 6.8 & \multirow{3}*{\revone{\cmark}}\\
         TinyViT-11M & 11.0 & 2.0 & 81.5 & 12.0 & \\
         \cellcolor{gray!30}{TinyViT-21M} & \cellcolor{gray!30}{21.0} & \cellcolor{gray!30}{4.3} & \cellcolor{gray!30}{83.1} & \cellcolor{gray!30}{24.3} & \\
         \hline
         DearKD-DeiT-Ti & 5 & \textbf{1.3} & 77.0 & 6.6 & \multirow{3}*{\revone{\cmark}} \\
         DearKD-DeiT-S & 22 & 4.6 & 82.8 & 25.7 & \\
         DearKD-DeiT-B & 87 & 17.6 & \textbf{84.4} & 100.2 & \\
         \hline
         CivT-Ti & 6 & - & 74.9 & - & \multirow{2}*{\revone{\cmark}}\\
         CivT-S & 22 & - & 82.0 & - & \\
         \hline
         Mini-DeiT-Ti & \textbf{3} & \textbf{1.3} & 72.8 & \textbf{5.4} & \multirow{6}*{\revone{\cmark}} \\
         Mini-DeiT-S & 11 & 4.7 & 80.7 & 19.7 & \\
         Mini-DeiT-B & 44 & 17.6 & 83.2 & 75.4 & \\
         Mini-Swin-T & 12 & 4.6 & 81.4 & 20.0 & \\
         Mini-Swin-S & 26 & 8.9 & 83.6 & 40.3 & \\
         Mini-Swin-B & 46 & 15.7 & 84.3 & 71.2 & \\
         \hline
         DynamicViT-LV-S/05 & 26.9 & 3.7 & 82.0 & 26.0 & \multirow{4}*{\revone{\cmark}}\\
         DynamicViT-LV-S/07 & 26.9 & 4.6 & 83.0 & 28.9 & \\
         DynamicViT-LV-M/07 & 57.1 & 8.5 & 83.8 & 57.1 & \\
         DynamicViT-LV-M/08 & 57.1 & 9.6 & 83.9 & 60.2 & \\
         \hline
    \end{tabular}
    \label{tab:results_on_imagenet1k_KD} 
\end{table}

Based on the reported results, it can be noticed that from the preliminary studies of Touvron et al. in 2021, the introduced DeiT architecture has been gradually enhanced and optimized. 
More in detail, this evolution can be noticed in further depth with the Manifold Distillation and DearKD strategies proposed in 2022.
These efficient techniques achieve an average accuracy improvement of $+1.5\%$ and $+1.7\%$, respectively, while keeping the number of trainable parameters and FLOPs constant.
Moreover, in the application scenario of extremely stringent computational constraints, the Mini-Deit-Ti model obtained with the MiniViT KD learning technique could be a valuable option with a very low EER and adequate classification performance.
However, we identify as the most efficient model between the one compared in Table~\ref{tab:results_on_imagenet1k_KD}, the TinyViT-21M student 
, which is able to achieve accurate estimations (Top-1 $=83.1\%$) with a limited EER equal to $24.3\%$.
Precisely, when comparing TinyViT-21M to similar models such as DearKD-DeiT-S, DynamicViT-LV-S/05, and DynamicViT-LV-S/07, no other student designs achieve superior classification scores with a restricted EER value.

\subsection{Results of Quantization strategies}
\label{subsec:results_of_Q}
In this section, we compare the estimation performances of general-purpose quantization strategies, which have been introduced in Section~\ref{sec:quantization} and formalized in Section~\ref{sec:quantization_background}.
As commonly reported in the reference papers, the different strategies are compared to the same architecture structure; precisely, we report the results obtained over ViT, DeiT, and Swin architectures, respectively, in Table~\ref{tab:results_on_imagenet1k_Q_ViT}, Table~\ref{tab:results_on_imagenet1k_Q_DeiT}, and Table~\ref{tab:results_on_imagenet1k_Q_Swin}.  

\begin{table}[t]
    \centering
    \footnotesize
    \caption{Quantitative comparison of \textbf{Q} strategies applied to  ViT models on ImageNet1K classification dataset. The best results are in bold, and the best (trade-off) efficient model is highlighted in gray.
    Values reported with $^*$ are estimated.}    
    \begin{tabular}{l | c c c c c }
         \multirow{2}*{Model (ViT)} & \multicolumn{2}{c}{\#Bit} & Size & Top-1 & EER \\
          & W & A & [MB] & [\%] & [\%] \\
         \hline
         Liu$_Q$-ViT-B & 6 & 6 & 64.8 & 75.2 & 19.2 \\
         Liu$_Q$-ViT-L & 6 & 6 & 231.6 & 75.5 & 44.4 \\
         Liu$_Q$-ViT-B & 8 & 8 & 86.5 & 76.9 & 25.6 \\
         Liu$_Q$-ViT-L & 8 & 8 & 306.4 & 76.4 & 58.9 \\
         \hline
         PTQ4ViT-ViT-S & 6 & 6 & 16.5 & 78.6 & 11.9 \\
         PTQ4ViT-ViT-B & 6 & 6 & 64.8 & 81.6 & 19.2 \\
         PTQ4ViT-ViT-S & 8 & 8 & 22.2 & 81.0 & 15.8 \\
         PTQ4ViT-ViT-B & 8 & 8 & 86.5 & 84.2 & 25.6 \\
         \hline
         APQ-ViT-ViT-T & 4 & 4 & \textbf{2.9} & 17.6 & \textbf{6.3} \\
         APQ-ViT-ViT-S & 4 & 4 & 11.1 & 47.9 & 7.9 \\
         APQ-ViT-ViT-B & 4 & 4 & 43.0 & 41.4 & 12.8 \\
         APQ-ViT-ViT-T & 8 & 4 & 4.1$^*$ & 38.6 & 9.9 \\
         APQ-ViT-ViT-S & 8 & 4 & 16.5$^*$ & 67.2 & 11.9\\
         APQ-ViT-ViT-B & 8 & 4 & 64.8$^*$ & 72.5 & 19.2 \\
         APQ-ViT-ViT-T & 4 & 8 & 4.1$^*$ & 59.4 & 18.4 \\
         APQ-ViT-ViT-S & 4 & 8 & 16.5$^*$& 72.3 & 11.9 \\
         APQ-ViT-ViT-B & 4 & 8 & 64.8$^*$& 72.6 & 19.2 \\
         APQ-ViT-ViT-T & 6 & 6 & 4.1 & 69.5 & 18.4 \\
         APQ-ViT-ViT-S & 6 & 6 & 16.5 & 79.1 & 11.9 \\
         APQ-ViT-ViT-B & 6 & 6 & 64.8 & 82.2 & 19.2 \\
         APQ-ViT-ViT-T & 8 & 8 & 5.4 & 74.8 & 13.3\\
         APQ-ViT-ViT-S & 8 & 8 & 22.2 & 81.2 & 15.8 \\
         APQ-ViT-ViT-B & 8 & 8 & 86.5 & \textbf{84.3} & 25.6 \\
         \hline
         NoisyQuant-PTQ4ViT-ViT-S & 6 & 6 & 16.5 & 78.6 & 11.9\\
         \cellcolor{gray!30}{NoisyQuant-PTQ4ViT-ViT-B} & \cellcolor{gray!30}{6} & \cellcolor{gray!30}{6} & \cellcolor{gray!30}{64.8} & \cellcolor{gray!30}{82.3} & \cellcolor{gray!30}{19.2} \\
         NoisyQuant-PTQ4ViT-ViT-S & 8 & 8 & 22.2 & 81.1 & 15.8 \\
         NoisyQuant-PTQ4ViT-ViT-B & 8 & 8 & 86.5 & 84.2 & 25.6 \\
         \hline
    \end{tabular}
    \label{tab:results_on_imagenet1k_Q_ViT} 
\end{table}

Differently from previous studies, these tables report the bit-width (\#Bit) in which the models are compressed for both weights (W) and activation functions (A), as well as the size (Size) of the model after the quantization. 
This choice is due to the fundamental effect of the Q efficient strategy, which tries to preserve the estimation performances of originally trained models with a lower data precision, i.e., compressing their data from 32-bit to 8/6/4-bit.
As a result, after the Q methodology, the number of trainable parameters (\#Pram.) will remain constant, but the data precision of each neuron, and hence the total size of the model, will be reduced.

The first performed analysis regards the optimal bit-width in order to generate efficient and accurate models. 
Based on the data presented in the multiple tables, it can be noticed that an extreme quantization, i.e., 4-bit precision, results in poor accuracy.
More in detail, the APQ-ViT quantization strategy applied to the ViT-T and DeiT-Ti architectures, which have the smaller model's sizes, produce a Top-1 accuracy equal to $17.6\%$ and $47.9\%$, respectively. 
In contrast, the same architectures with higher (8/6) bit-width are able to achieve adequate estimation performances with a Top-1 up to $74.8\%$ and $72.0\%$ for the ViT-T and DeiT-Ti structures respectively, while still maintaining a restricted model size.
Furthermore, by observing the heavier Swin models, as shown in Table~\ref{tab:results_on_imagenet1k_Q_Swin}, it can be noticed that even with an extreme quantization (4-bit), those architectures can guarantee good performance, with a Top-1 accuracy of $77.1\%$ and $76.5\%$ for APQ-ViT-Swin-S and APQ-ViT-Swin-B respectively.
To summarize, severe quantization strategies applied on small models often result in significant performance degradation; consequently, it would be preferable to use 6/8-bit approaches in the case of shallower models while using 4-bit compression for deeper architectures.
However, an extreme quantization strategy could benefit in applications where computational requirements are a bottleneck for ViT inference or on specific hardware, such as in the case of Google Coral TPUs\footnote{\texttt{https://coral.ai/products/}}. 

\begin{table}[t]
    \centering
    \footnotesize
    \caption{Quantitative comparison of \textbf{Q} strategies applied to DeiT models on ImageNet1K classification dataset. The best results are in bold, and the best (trade-off) efficient model is highlighted in gray.
    Values reported with $^*$ are estimated.}   
    \begin{tabular}{l | c c c c c }
         \multirow{2}*{Model (DeiT)} & \multicolumn{2}{c}{\#Bit} & Size & Top-1 & EER \\
          & W & A & [MB] & [\%] & [\%] \\
         \hline
         Liu$_Q$-DeiT-B & 4 & 4 & 43.6 & 75.9 & 12.8 \\
         Liu$_Q$-DeiT-S & 6 & 6 & 16.6 & 75.1 & 11.9 \\
         Liu$_Q$-DeiT-B & 6 & 6 & 64.3 & 77.5 & 19.1 \\
         Liu$_Q$-DeiT-S & 8 & 8 & 22.2 & 78.1 & 15.8\\
         Liu$_Q$-DeiT-B & 8 & 8 & 86.8 & 81.3 & 25.6 \\
         \hline
         PTQ4ViT-DeiT-S & 6 & 6 & 16.6 & 76.3 & 11.9 \\
         PTQ4ViT-DeiT-B & 6 & 6 & 64.3 & 80.2 & 19.1 \\
         PTQ4ViT-DeiT-S & 8 & 8 & 22.2 & 79.5 & 15.8\\
         PTQ4ViT-DeiT-B & 8 & 8 & 86.8 & 81.5 & 25.6\\
         \hline
         APQ-ViT-DeiT-Ti & 4 & 4 & \textbf{2.5} & 47.9 & \textbf{6.6} \\
         APQ-ViT-DeiT-S & 4 & 4 & 11.0 & 43.5 & 7.9 \\
         APQ-ViT-DeiT-B & 4 & 4 & 43.6 & 67.5 & 12.8 \\
         APQ-ViT-DeiT-Ti & 8 & 4 & 3.7$^*$ & 56.3 & 9.9 \\
         APQ-ViT-DeiT-S & 8 & 4 & 16.6$^*$ & 41.3 & 11.9 \\
         APQ-ViT-DeiT-B & 8 & 4 & 64.3$^*$ & 71.7 & 19.1 \\
         APQ-ViT-DeiT-Ti & 4 & 8 & 3.7$^*$ & 66.7 & 9.9 \\
         APQ-ViT-DeiT-S & 4 & 8 & 16.6$^*$ & 77.1 & 11.9\\
         APQ-ViT-DeiT-B & 4 & 8 & 64.3$^*$ & 79.5 & 19.1\\
         APQ-ViT-DeiT-Ti & 6 & 6 & 3.7 & 70.5 & 9.9\\
         APQ-ViT-DeiT-S & 6 & 6 & 16.6 & 77.8 & 11.9 \\
         APQ-ViT-DeiT-B & 6 & 6 & 64.3 & 80.4 & 19.1\\
         APQ-ViT-DeiT-Ti & 8 & 8 & 5.0 & 72.0 & 10.1 \\
         APQ-ViT-DeiT-S & 8 & 8 & 22.2 & 79.8 & 15.8\\
         APQ-ViT-DeiT-B & 8 & 8 & 86.8 & \textbf{81.7} & 25.6 \\
         \hline
         NoisyQuant-PTQ4ViT-DeiT-S & 6 & 6 & 16.6 & 77.4 & 11.9 \\
         \cellcolor{gray!30}{NoisyQuant-PTQ4ViT-DeiT-B} & \cellcolor{gray!30}{6} & \cellcolor{gray!30}{6} & \cellcolor{gray!30}{64.3} & \cellcolor{gray!30}{80.7} & \cellcolor{gray!30}{19.1} \\
         NoisyQuant-PTQ4ViT-DeiT-S & 8 & 8 & 22.2 & 79.5 & 15.8\\
         NoisyQuant-PTQ4ViT-DeiT-B & 8 & 8 & 86.8 & 81.4 & 25.6 \\
         \hline
    \end{tabular}
    \label{tab:results_on_imagenet1k_Q_DeiT} 
\end{table}

Finally, in order to determine the Paretian efficiency between compared methodologies and architectures, we will first identify the best trade-off solution for each given table and then establish the ideal optimal quantization strategy.
Based on the reported data, we identify the NoisyQuant-PTQ4ViT quantization approach as the best trade-off option among compared ones. 
However, we emphasize that also APQ-ViT performs quite similarly to NoisyQuant-PTQ4ViT and may be considered a suitable alternative.
Moreover, similarly to previous study, in order to ensure a Top-1 accuracy greater than $80.0\%$, we identify the ViT-B, DeiT-B, and Swin-B architectures as the best trade-off solution for each compared table adopting the 6-bit NoisyQuant-PTQ4Vi quantization strategy.
However, we determined the NoisyQuant-PTQ4ViT-Swin-B methodology as the most efficient model among all analyzed solutions. 
Specifically, the latter architecture achieves a Top-1 accuracy equal to $84.7\%$ with a model size of $66.0$MB; in contrast, the ViT-B and DeiT-B variants obtain inferior classification performances of $-2.3\%$ and $-4.0\%$ with almost equal EER values and model sizes.

\begin{table}[t]
    \centering
    \footnotesize
    \caption{Quantitative comparison of \textbf{Q} strategies applied to Swin models on ImageNet1K classification dataset. The best results are in bold, and the best (trade-off) efficient model is highlighted in gray.} 
    \begin{tabular}{l | c c c c c }
         \multirow{2}*{Model} & \multicolumn{2}{c}{\#Bit} & Size & Top-1 & EER \\
          & W & A & [MB] & [\%] & [\%] \\
         \hline
         PTQ4ViT-Swin-T & 6 & 6 & \textbf{21.7} & 80.5 & 12.5 \\
         PTQ4ViT-Swin-S & 6 & 6 & 37.5 & 82.4 & 15.0 \\
         PTQ4ViT-Swin-B & 6 & 6 & 66.0 & 84.0 & 19.4 \\
         PTQ4ViT-Swin-T & 8 & 8 & 29.0 & 81.2 & 16.9 \\
         PTQ4ViT-Swin-S & 8 & 8 & 50.0 & 83.1 & 20.1 \\
         PTQ4ViT-Swin-B & 8 & 8 & 88.0 & 85.1 & 25.9 \\
         \hline
         APQ-ViT-Swin-S & 4 & 4 & 25.0 & 77.1 & \textbf{10.0}\\
         APQ-ViT-Swin-B & 4 & 4 & 44.0 & 76.5 & 12.9 \\
         APQ-ViT-Swin-S & 8 & 4 & 37.5 & 80.6 & 15.0 \\
         APQ-ViT-Swin-B & 8 & 4 & 66.0 & 82.0& 19.4 \\
         APQ-ViT-Swin-S & 4 & 8 & 37.5 & 80.6 & 15.0 \\
         APQ-ViT-Swin-B & 4 & 8 & 66.0 & 81.9 & 19.4 \\
         APQ-ViT-Swin-S & 6 & 6 & 37.5 & 82.7 & 15.0\\
         APQ-ViT-Swin-B & 6 & 6 & 66.0 & 84.2 & 19.4 \\
         APQ-ViT-Swin-S & 8 & 8 & 50.0 & 83.2 & 20.1 \\
         APQ-ViT-Swin-B & 8 & 8 & 88.0 & \textbf{85.2} & 25.9 \\
         \hline
         NoisyQuant-PTQ4ViT-Swin-T & 6 & 6 & \textbf{21.7} & 80.5 & 12.5\\
         NoisyQuant-PTQ4ViT-Swin-S & 6 & 6 & 37.5 & 82.8& 15.0 \\
         \cellcolor{gray!30}{NoisyQuant-PTQ4ViT-Swin-B} & \cellcolor{gray!30}{6} & \cellcolor{gray!30}{6} & \cellcolor{gray!30}{66.0} & \cellcolor{gray!30}{84.7} & \cellcolor{gray!30}{19.4} \\
         NoisyQuant-PTQ4ViT-Swin-T & 8 & 8 & 29.0 & 81.2& 16.9\\
         NoisyQuant-PTQ4ViT-Swin-S & 8 & 8 & 50.0 & 83.1& 20.1 \\
         NoisyQuant-PTQ4ViT-Swin-B & 8 & 8 & 88.0 & \textbf{85.2} & 25.9\\
         \hline
    \end{tabular}
    \label{tab:results_on_imagenet1k_Q_Swin} 
\end{table}

\subsection{Overall classification performances}
\label{subsec:results_of_CAPKDQ_calass}
In this section, the best efficient strategies identified among the CA, P, KD, and Q categories are compared.
Moreover, in order to provide a comprehensive and broad overview on limitations and estimation performances, we report in Table~\ref{tab:results_on_imagenet1k_CAPKDQ} an inter-category comparison.
Precisely, we analyze each strategy under all the evaluation metrics used in the previous analysis, i.e., bit-width (\#Bit), model size, number of parameters (\#Par.), FLOPs, accuracy (Top-1) and the EER (EER$_{catg}$) value computed in the respective category.
Furthermore, by employing in the Equation~\ref{eq:eer_metric} the set of metrics $i\in \{\#Bit, Size, \#Par., FLOPs \}$, we generate a category-independent EER value (EER$_{all}$) that takes into account all the compared metrics.

\begin{table*}[t]
    \centering
    \footnotesize
    \caption{Quantitative comparison between best \textbf{CA}, \textbf{P}, \textbf{KD}, and \textbf{Q} identified strategies on ImageNet1K classification dataset. The best results are in bold, and the best (trade-off) efficient model is highlighted in gray.
    Values reported with $^*$ are estimated.
    EER$_{catg}$ corresponds to the EER value computed in the respective category, while EER$_{all}$ corresponds to the EER value computed over all the reported metrics.} 
    \begin{tabular}{ l | l | c c c c c c c | c }
          \multirow{2}*{Category} &  \multirow{2}*{Model} & \multicolumn{2}{c}{\#Bit} & Size & \#Par. & FLOPs & Top-1 & EER$_{catg}$ & EER$_{all}$ \\
           & & W & A & [MB] & [M] & [G] & [\%] & [\%]  & [\%] \\
          \hline
          \cellcolor{gray!30}{CA} & \cellcolor{gray!30}{Castling-MViTv2-T} & \cellcolor{gray!30}{32} & \cellcolor{gray!30}{32} & \cellcolor{gray!30}{96.4} & \cellcolor{gray!30}{24.1} & \cellcolor{gray!30}{4.5} & \cellcolor{gray!30}{84.1} & \cellcolor{gray!30}{26.7} & \cellcolor{gray!30}{45.6} \\
          \hline
          P + KD & DynamicViT-LV-S/07 & 32 & 32 & 107.6 & 26.9 & 4.6 & 83.0 & 28.9 & 47.4 \\
          \hline
          KD & TinyViT-21M & 32 & 32 & 84.0 & \textbf{21.0} & \textbf{4.3} & 83.1 & 24.3 & \textbf{43.5} \\
          \hline
          Q & NoisyQuant-PTQ4ViT-Swin-B & \textbf{6} & \textbf{6} & \textbf{66.0} & 88.0 & 15.4 & \textbf{84.7} & \textbf{19.4} & 56.5 \\
          \hline
    \end{tabular}
    \label{tab:results_on_imagenet1k_CAPKDQ} 
\end{table*}

Based on the reported results, it is possible to notice that DynamicViT-LV/07 and TinyViT-21M strategies are both obtained via a KD learning technique. 
More in detail, DynamicViT-LV/07 is a pruning strategy that leverages the use of KD to dynamically and progressively shrink the pruned-student model in order to minimize the performance reduction caused by the model's sparsification. 
Even so, TinyViT-21M (student) model achieves comparable classification performances ($\simeq83\%$) with a smaller and lighter model, i.e., $-5.9$M of trainable parameters, $-0.3$G FLOPs and $-23.6$MB.
Moreover, thanks to the memory-efficient KD strategy introduced by Wu et al., the TinyViT-21M strategy obtains the lower EER$_{all}$ value equal to $43.5\%$ when computed across all the reported evaluation metrics.

In contrast, when TinyViT-21M is compared with the NoisyQuant-PTQ4ViT-Swin-B approach, it can be noticed that having a greater number of FLOPs computed with a lower precision (6-bit) can be beneficial for the model size and the estimation performances.
In comparison to TinyViT-21M, NoisyQuant-PTQ4ViT-Swin-B achieves the maximum Top-1 accuracy of $84.7\%$, with a tiny model of $66.0$MB with respect to TinyViT-21M ($84.0$MB).
Precisely, the quantized strategy achieves a Top-1 boost of $+1.6\%$ with a model that is $-21.4\%$ smaller.
However, as also underlined by Liu et al. (EcoFormer), the use of specific CUDA kernels or optimized interpreters may be necessary to maximize the effective performance of these quantization-based algorithms. 
Therefore, the NoisyQuant-PTQ4ViT-Swin-B may be an effective and efficient strategy for specific hardware setups.

Finally, the Paretian optimality between Top-1 accuracy and EER$_{all}$ is obtained by the CA strategy with the Castling-MViTv2-T model.
As shown in Table~\ref{tab:results_on_imagenet1k_CAPKDQ}, the latter architecture achieves optimal estimation performances (Top-1 $=84.1\%$) while maintaining a suitable EER$_{all}$.
Precisely, when compared with TinyViT-21M, the Castling-MViTv2-T model achieves an accuracy boost of $+1.1\%$ at the expense of an EER$_{all}$ increment of $2.1\%$; differently, when compared with NoisyQuant-PTQ4ViT-Swin-B, the accuracy is reduced by $-0.6\%$, and the EER$_{all}$ is increased by $+10.9\%$.
However, as can be seen from the reported results, the Castling-MViTv2-T model does not excel in any of the reported metrics (bold in Table~\ref{tab:results_on_imagenet1k_CAPKDQ}); this is because the objective of this final analysis is not to identify the best strategy for making ViTs efficient but rather to identify the model that best balances efficiency and classification performances in general-purpose settings.

\section{Discussions and Conclusion}
\label{sec:conclusion_and_discussions}
In this paper, we survey the literature on efficient ViT strategies, one of the main bottlenecks of such architectures.
The available techniques were classified into four categories: compact architecture (CA), pruning (P), knowledge distillation (KD), and quantization (Q). 
Moreover, we formalize their algorithms (Section~\ref{sec:general_background}), we review proposed techniques in order to highlight their strengths and weaknesses (Section~\ref{sec:eff_ViT}), and finally, we compare the model/strategy performances over well-known benchmark datasets (Section~\ref{sec:results_and_applications}) in order to determine the best trade-off strategy between estimation performances and efficiency.
Furthermore, we introduce a novel evaluation metric named Efficient Error Rate (EER), which is used to give a general overview of the efficiency performances of reviewed models in comparison to a predefined and fixed non-efficient baseline solution.

In conclusion, based on what has been previously determined and analyzed so far, this last section will focus on open challenges and promising research directions that may improve the efficiency of general-purpose ViT models.
The reviewed works are only one of the first steps toward developing effective and efficient solutions, leaving much room for future improvement.
Therefore, we identified three potential major limitations, which are following discussed: application/tasks, evaluation metrics, and merging strategy.

\vspace{1.0em}
\noindent
\textbf{Application/Task:} \revtwo{As observed from the preceding analysis, many of the compared efficient approaches solely rely on testing their respective performances on a well-known classification task; in contrast, in CA optimizations, a variety of scenarios have been tested.
Therefore, exploring these efficient architectures in different or more complex tasks is necessary.
Furthermore, whether these architectures can generalize to other tasks or be robust to adversarial attacks are fundamental notions that still need to be investigated and explained.}

\vspace{1.0em}
\noindent
\textbf{Evaluation metrics:} Based on the tables reported in Section~\ref{sec:results_and_applications} and on the reviewed works, the metrics used to determine the quality of proposed solutions, such as accuracy (Top-1/AP/mIOU), FLOPs, and number of trainable parameters, are evaluation metrics and architecture's values commonly used for general-purpose tasks. 
Consequently, since the development of efficient solutions aims to determine the best trade-off between accuracy and efficiency in order to apply ViTs in real-world scenarios, the development of specific benchmark tests based on resource-constrained devices could aid in comparing proposed solutions in both intra and inter-categories. 
More in detail, these metrics could focus on assessing the degradation effect caused by optimization strategies; this is because lowering the model's computational cost and associated resource demands typically comes at the expense of a performance decrease.

\vspace{1.0em}
\noindent
\textbf{Merging strategies:} The attention mechanism has played a crucial role in the development of accurate computer vision systems; however, the reduction/linearization of its computational cost has not been without performance consequences.
According to the summary reported in Table~\ref{tab:summary_survey}, it is also visible that compared strategies mainly focus on determining the most efficient solution only focusing on a specific technique, i.e., CA, P, KD, or Q. 
In contrast, the ability to combine multiple techniques, as exemplified in DynamicViT by Rao et al., might successfully limit the accuracy loss of shallower architecture with respect to deeper one.
As a result, since each efficient strategy leads to a reduction of ViT computational requirements, studying solutions that employ multiple efficient methodologies and determining the optimal way these solutions can be integrated may be an effective pathway for developing future efficient methodologies.

\printbibliography


\begin{IEEEbiography}[{\includegraphics[width=1in,height=1.25in,clip,keepaspectratio]{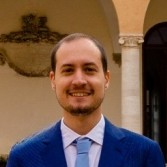}}]{LORENZO PAPA} (Student Member, IEEE) is a Ph.D. student in Computer Science Engineering. 
He collaborates with AlcorLab at the Department of Computer, Control, and Management Engineering, Sapienza University of Rome, Italy.
He is a Visiting Researcher at the School of Electrical and Information Engineering, Faculty of Engineering, The University of Sydney, Australia. 
He received the B.S. degree in Computer and Automation Engineering and the M.S. degree in Artificial Intelligence and Robotics from Sapienza University of Rome, Italy, in 2019 and 2021, respectively.
His main research interests are Deep Learning, Computer Vision, and Cyber Security.
\end{IEEEbiography}

\begin{IEEEbiography}[{\includegraphics[width=1in,height=1.25in,clip,keepaspectratio]{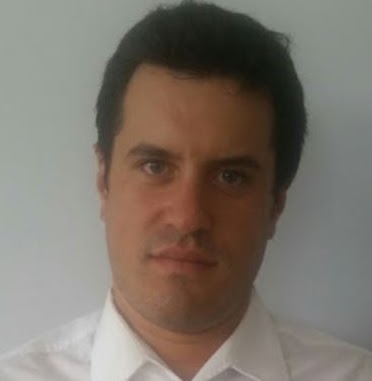}}]{PAOLO RUSSO} is an Assistant Researcher at AlcorLab in DIAG department, University of Rome Sapienza, Italy. 
He received the B.S. degree in Telecommunication Engineering from Università degli studi di Cassino, Italy, in 2008, and the M.S. degree in Artificial Intelligence and Robotics from University of Rome La Sapienza, Italy, in 2016. 
He received Ph.D. degree in Computer Science from University of Rome La Sapienza in 2020.
From 2018 to 2019, he has been a researcher at Italian Institute of Technology (IIT) in Tourin, Italy.
His main research interests are Deep Learning, Computer Vision, Generative Adversarial Networks, and Reinforcement Learning.
\end{IEEEbiography}

\begin{IEEEbiography}[{\includegraphics[width=1in,height=1.25in,clip,keepaspectratio]{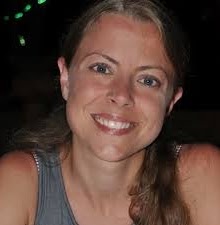}}]{IRENE AMERINI} (M’17) received Ph.D. degree in computer engineering, multimedia, and telecommunication from the University of Florence, Italy, in 2010. 
She is currently Associate Professor with the Department of Computer, Control, and Management Engineering A. Ruberti, Sapienza University of Rome, Italy.  
Her main research activities include digital image processing, computer vision multimedia forensics. She is a member of the IEEE Information Forensics and Security Technical Committee and the EURASIP TAC Biometrics, Data Forensics, and Security, and the IAPR TC6 - Computational Forensics Committee.
\end{IEEEbiography}

\begin{IEEEbiography}[{\includegraphics[width=1in,height=1.25in,clip,keepaspectratio]{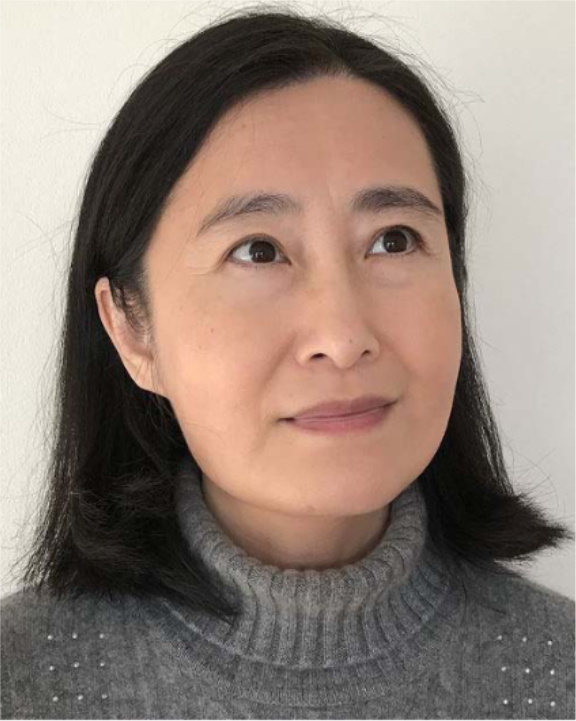}}]{LUPING ZHOU} (Senior Member, IEEE) received the Ph.D. degree from Australian National University. 
She is currently an Associate Professor with the School of Electrical and Information Engineering, The University of Sydney, Australia. 
Her research interests include medical image analysis, machine learning, and computer vision. She was a recipient of Australian Research Council Discovery Early Career Researcher (DECRA) Award in 2015.
\end{IEEEbiography}

\vfill

\newpage
\section{Supplementary material}
\label{sec:supp_material}
The supplementary material is organized into four subsections: Section~\ref{subsec:dynamic_inference} reviews state-of-the-art efficient deep learning solutions for dynamic inference methodologies, Section~\ref{subsec:ERR_metric} proposes an in-depth analysis of the applicability and flexibility of the proposed EER metric and its weighted variant.
Finally, Section~\ref{subsec:coco_adek} compares CA methodologies over object detection, instance segmentation, and semantic segmentation tasks, and Section~\ref{subsec:graphs} reports a graphical representation of CA, KD, and P model’s distributions regarding accuracy-EER values.

\subsection{Dynamic Inference}
\label{subsec:dynamic_inference}

\revone{Dynamic inference techniques are a family of frameworks that can adjust their architecture in response to different inputs/features. 
Those methodologies are usually based on adaptive strategies designed to lower the computational cost and inference timings of a general neural network model.
Therefore, in this section, we report an in-depth analysis of newly proposed works. 
However, we would like to underline that such methodologies are not only limited to pruning strategies like~\cite{rao2021dynamicvit, rao2023dynamic} but also integrated into CA design such as~\cite{koohpayegani2022sima}.

Yin et al.~\cite{yin2022vit} (2022) propose A-ViT, an input-dependent adaptive inference mechanism that is able to dynamically reduce the inference cost of ViT architectures without adding extra parameters or computations to the overall architecture.
More in detail, the authors add a halting probability neuron to each embedded patch, which is used to compute the halting score of the token; subsequently, if the halting condition is satisfied, the token is discarded (pruned), and the successive transformer block will only receive information from active tokens.
This strategy will steadily reduce the overall model's computation since only discriminative (active) tokens, i.e., the ones that are informative for the task, are processed by successive transformer blocks, resulting in fast inference performances.

Similarly, Kong et al.~\cite{kong2022spvit} (2022) introduce SPViT, a training strategy based on a latency-aware soft token pruning framework that is able to reduce the overall computation of vanilla ViT. 
However, differently to A-ViT, SPViT incorporates the less informative tokens identified by a selector module 
into a package token rather than discarding them completely; this approach is motivated by the trade-off accuracy-inference requirements needed to infer on specific edge devices.

Unlike pruning algorithms, Bolya et al.~\cite{bolya2023tome} (2023) introduce a Token Merging (ToMe) strategy which is able to merge redundant tokens into pretrained ViT architectures, enhancing the throughput without retraining the model. 
Precisely, each transformer layer is reduced by a specific factor (value) to identify the best speed-accuracy trade-off since fewer tokens mean lower accuracy but higher inference performances.
ToMe is based on the Token Similarity algorithm, which establishes that two tokens are defined similar (and so can be merged) if the distance between their features is small. 
Finally, the authors also provide a proportional self-attention module needed to maintain the original input patch representation.
Specifically, by defining with $s$ a vector containing the size of each token, the introduced attention function can be formulated as reported in Equation~\ref{eq:ToMe}.
\begin{equation}
    \label{eq:ToMe}
    A_{ToMe} = Softmax \bigg ( \frac{Q\cdot K^T}{\sqrt{d_k}} + log(s) \bigg )
\end{equation}

}

\subsection{Applicability and flexibility of the ERR metric}
\label{subsec:ERR_metric}

\revone{In Section~\ref{sec:results_and_applications}, we introduce the Efficient Error Rate (EER) metric, centered on a specific architecture and on a restricted set of metrics required to evaluate the models analyzed in Section~\ref{sec:eff_ViT}.
However, in this section, we would like to highlight the adaptability and flexibility of the introduced EER metric across diverse real-world scenarios with various constraints.

The EER metric has been designed to combine several non-homogeneous (values with distinct units of measurement) parameters of efficient architectures into a single evaluation metric, i.e., under a single value.
This choice is mainly due to the fact that in resource-constrained circumstances, each application/task has its own restriction, and it is uncommon to have a well-known unique evaluation metric to have a fast and broad overview of the developed architecture.
Moreover, the estimation accuracy value commonly employed in a variety of applications may not be the only factor used to determine the overall quality of a model; this is especially true when a multitude of constraints (model size, latency, precision data type) have to be considered in order to design a lightweight neural network to infer on a specific embedded device. 
Therefore, as opposite to the accuracy, we propose an error metric, i.e., a metric in which the lower its value, the better, that is capable of simultaneously taking into account all the parameters/set of efficiency metrics that have to be minimized in order to provide a general overview of the efficient capability of the developed model with respect to the reference constraints/task requirements.

In the following are reported the needed steps in order to use the metric:
\begin{enumerate}
    \item \textbf{Parameter's selection:} The first step is to identify the parameters/efficiency metrics ($i\in\{\#Param, FLOPs, ...\}$) and their respective values ($R_i$) that we have to minimize (upper bounds); they can simultaneously influence the development and the behavior of a neural network. 
    For example, in the case of embedded devices and microprocessors, the limited computational resources available make the size of the model a critical feature. 
    Similarly, in real-time applications, the inference time is crucial for the model development, while in green scenarios, variables such as energy consumption and the model's carbon footprint have also been considered. 
    \item \textbf{EER Computation:} Subsequently, once each reference value for each parameter has been identified, instead of considering each of the previous metrics independently, we can compute the EER metric as reported in Equation~\ref{eq:eer_metric}.
    Consequently, based on the obtained EER value, we can have an overall understanding of the efficiency performances of the designed model with respect to a reference one.
\end{enumerate}

Based on Equation~\ref{eq:eer_metric}, each parameter has the same impact on the EER calculation. 
However, in some circumstances, some parameters may hold greater relevance; under these settings, the metric can be weighted by assigning a scaling factor ($\delta$) for each parameter, such that their sum equals one.
We can formalize the weighted EER metric as reported in Equation~\ref{eq:eer_metric_weigthed}.
\begin{equation}
    \label{eq:eer_metric_weigthed}
    EER = \frac{1}{||i||} \cdot \sum_{i} \delta_i \cdot \bigg( \frac{M_i}{R_i} \bigg)
\end{equation}

Finally, we believe that combining the proposed EER metric with the well-known accuracy measurements would aid in developing future efficient structures, in order to determine the Paretian optimality between accuracy and the EER in the reference application scenario.}

\subsection{CA strategy COCO and ADE20K datasets}
\label{subsec:coco_adek}

\freespace{In this subsection we compare the CA performances on two additional datasets, i.e., the COCO~\cite{lin2014microsoft} object detection and instance segmentation dataset, which is composed of $118$K training (\texttt{train2017}) and $5$K validation (\texttt{val2017}) images at a resolution of $800\times~1,333$ pixels; and the ADE20K~\cite{zhou2017scene} semantic segmentation datasets, with $20$K, $2$K, and $3$K images for training, validation, and testing, respectively resized at a resolution of $512 \times 512$ pixels.}

\vspace{1.0em}
\noindent
\freespace{\textbf{Object Detection and Instance Segmentation:} 
In order to provide a more general overview of CA techniques, we briefly evaluate and compare the efficient CA solutions also on the object detection and instance segmentation task over the COCO dataset.
The results, reported by employing the Mask R-CNN~\cite{he2017mask} (1x) as decoder, are presented in Table~\ref{tab:results_CA_on_coco_objdect}.
More in detail, to show the performance of the proposed models, some of the reported studies focus on a variety of decoders such as RetinaNet~\cite{lin2017focal} (1x) and Cascade Mask R-CNN~\cite{cai2018cascade}  
However, we only reported in Table~\ref{tab:results_CA_on_coco_objdect} the results for Mask R-CNN (1x) architecture since it is the most widely used structure and best trade-off between shallow and deep decoders\footnote{RetinaNet (1x) and the Cascade Mask R-CNN have $-10.0$M and $+33.0$M of trainable parameter with respect to the Mask R-CNN (1x) respectively.}. 
Consequently, based on the reported values, we conclude that the MViTv2-T with the Mask R-CNN 1x as decoder provides the best encoder-decoder trade-off between estimate performances and EER values (only for the encoder) as shown in Table~\ref{tab:results_CA_on_coco_objdect} and Table~\ref{tab:results_on_imagenet1k_CA}.
More in detail, the MViTv2-T structure achieves comparable estimation performances, i.e., a decrement of $-3.6$ and $-2.4$, respectively, for the AP$^{box}$ and AP$^{mask}$ metrics, with respect to the more accurate MViTv2-L configuration with a structure which is $\times5.4$ smaller; while achieving a boost of $+11.5$ and $+8.7$ over the previous evaluation metrics, with respect to the lighter PVT-Tiny model.}

\begin{table}[t]
    \centering
    \footnotesize
    \caption{\freespace{Quantitative comparison of \textbf{CA} models on COCO (\texttt{val2017}) object detection and instance segmentation dataset. The best results are in bold, and the best (trade-off) efficient model is highlighted in gray.}}
    \begin{tabular}{l | c c c }
         \multirow{2}*{Encoder} & \#Par. & \multirow{2}*{AP$^{box}$} & \multirow{2}*{AP$^{mask}$} \\
          & [M] & & \\
         \hline
         PVT-Tiny & \textbf{32.9} & 36.7 & 35.1 \\
         PVT-Small & 44.1 & 40.4 & 37.8 \\
         PVT-Medium & 63.9 & 42.0 & 39.0 \\
         PVT-Large & 81.0 & 42.9 & 39.5 \\
         \hline
         PoolFormer-S12 & 31.6 & 37.3 & 34.6 \\
         PoolFormer-S24 & 41.3 & 41.0 & 37.0 \\
         PoolFormer-S36 & 50.5 & 41.0 & 37.7 \\
         \hline
         PVTv2-B2-LiSRA & 42.2 & 44.1 & 40.5 \\
         \hline
         \cellcolor{gray!30}{MViTv2-T} & \cellcolor{gray!30}{44} & \cellcolor{gray!30}{48.2} & \cellcolor{gray!30}{43.8} \\
         MViTv2-S & 54 & 49.9 & 45.1 \\
         MViTv2-B & 71 & 51.0 & 45.7 \\
         MViTv2-L & 238 & \textbf{51.8} & \textbf{46.2} \\
         \hline
         XCiT-T12/16 $\rightarrow$ SimA & 44.3 & 44.8 & 40.3 \\
         \hline
         Ortho-S & 44 & 47.0 & 42.5 \\
         Ortho-B & 69 & 48.3 & 43.3 \\
         \hline
         \end{tabular} 
    \label{tab:results_CA_on_coco_objdect}
\end{table}

\vspace{1.0em}
\noindent
\freespace{\textbf{Semantic Segmentation:} Similarly to the previous study, in this subsection, we compare the efficient CA proposed solutions on the semantic segmentation task performed over the ADE20K dataset.
The results, only for the available models, are reported in Table~\ref{tab:results_CA_on_ade20k}.
Based on the reported architectures, we can notice that similar to the object detection task, different decoders such as Semantic FPN~\cite{kirillov2019panoptic} and UperNet~\cite{xiao2018unified} structures have been employed.
These decoders significantly impact the number of trainable parameters of the compared efficient architectures since UperNet counts $+28$M with respect to Semantic FPN.
However, the higher number of trainable parameters is not justified by a noticeable mIoU improvement. 
The latter assumption can be justified by taking into account Ortho-S and Ortho-B models, which share the same encoder with different decoders; in these settings, the performance boost achieved by the heavier decoder (UperNet) with respect to Semantic FPN over the two encoders is equal to $+0.3\%$ and $+0.8\%$ respectively. 

Finally, we identify the best efficient encoder-decoder architecture, the Ortho-S backbone combined with the Semantic FPN decoder; the overall architecture achieves, in fact, remarkable estimation performances (mIoU$=48.2$) with a constrained number of trainable parameters ($28$M).}

\begin{table}[h]
    \centering
    \footnotesize
    \caption{\freespace{Quantitative comparison of \textbf{CA} models on ADE20K semantic segmentation dataset. 
    The best results are in bold, and the best (trade-off) efficient model is highlighted in gray.}}
    \begin{tabular}{l l | c c c }
         \multirow{2}*{Encoder} & \multirow{2}*{Decoder} &  \#Par. & mIOU  \\
         & & [M] & [\%] \\
         \hline
         PVT-Tiny & Semantic FPN & 17.0 & 35.7 & \\
         PVT-Small & Semantic FPN & 28.2 & 39.8 & \\
         PVT-Medium & Semantic FPN & 48.0 & 41.6 & \\
         PVT-Large & Semantic FPN & 65.1 & 42.1 & \\
         \hline
         Swin-T & UperNet & 60 & 46.1 & \\
         Swin-S & UperNet & 81 & 49.3 & \\
         Swin-B & UperNet & 121 & 51.6 & \\
         \hline
         SOFT-Small & UperNet & 54 & 46.2 & \\
         SOFT-Medium & UperNet & 76 & 48.0 & \\
         \hline
         PoolFormer-S12 & Semantic FPN & \textbf{15.7}\textbf{} & 37.2 & \\
         PoolFormer-S24 & Semantic FPN & 23.2 & 40.3 & \\
         PoolFormer-S36 & Semantic FPN & 34.6 & 42.0 & \\
         PoolFormer-M48 & Semantic FPN & 77.1 & 42.7 & \\
         \hline
         PVTv2-B2-LiSRA & Semantic FPN & 26.3 & 45.1 & \\
         \hline
         \cellcolor{gray!30}{Ortho-S} & \cellcolor{gray!30}{Semantic FPN} & \cellcolor{gray!30}{28} & \cellcolor{gray!30}{48.2} & \\
         Ortho-B & Semantic FPN & 53 & 49.0 & \\
         Ortho-S & UperNet & 54 & 48.5 & \\
         Ortho-B & UperNet & 81 & \textbf{49.8} & \\
         \hline
    \end{tabular}
    \label{tab:results_CA_on_ade20k}
\end{table}

\subsection{Architectures distributions regard to accuracy-EER values}
\label{subsec:graphs}

\begin{figure}[h]
    \centering
    \includegraphics[width=0.95\linewidth]{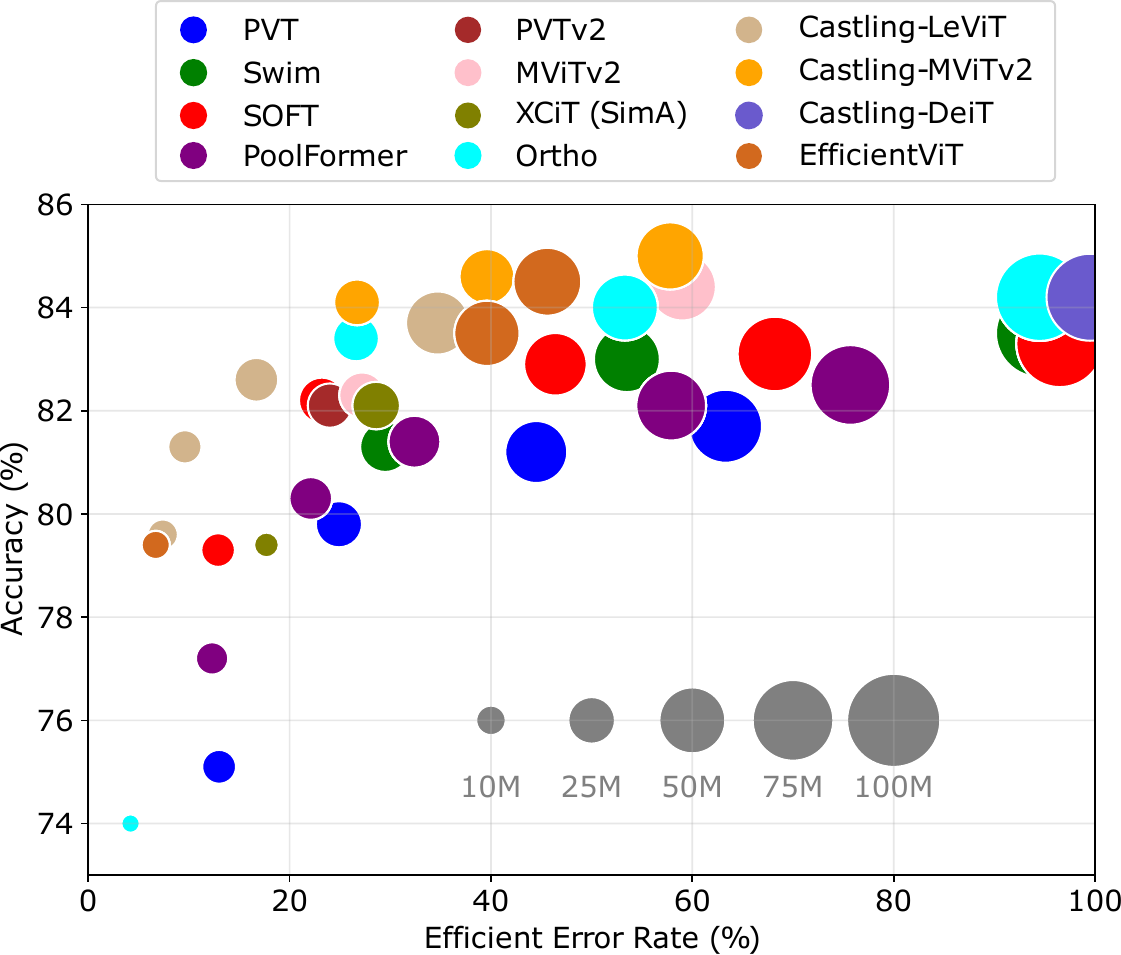}
    \caption{Graphical comparison of \textbf{CA} model Top-1 accuracies and EER values on the ImageNet-1K dataset. The area of bubbles corresponds to the number of trainable parameters (\#Param). The reference \#Param sizes (from 10M to 100M) are shown in gray in the bottom right corner.}
    \label{fig:CA_models_results}
\end{figure}

\begin{figure}[h]
    \centering
    \includegraphics[width=0.95\linewidth]{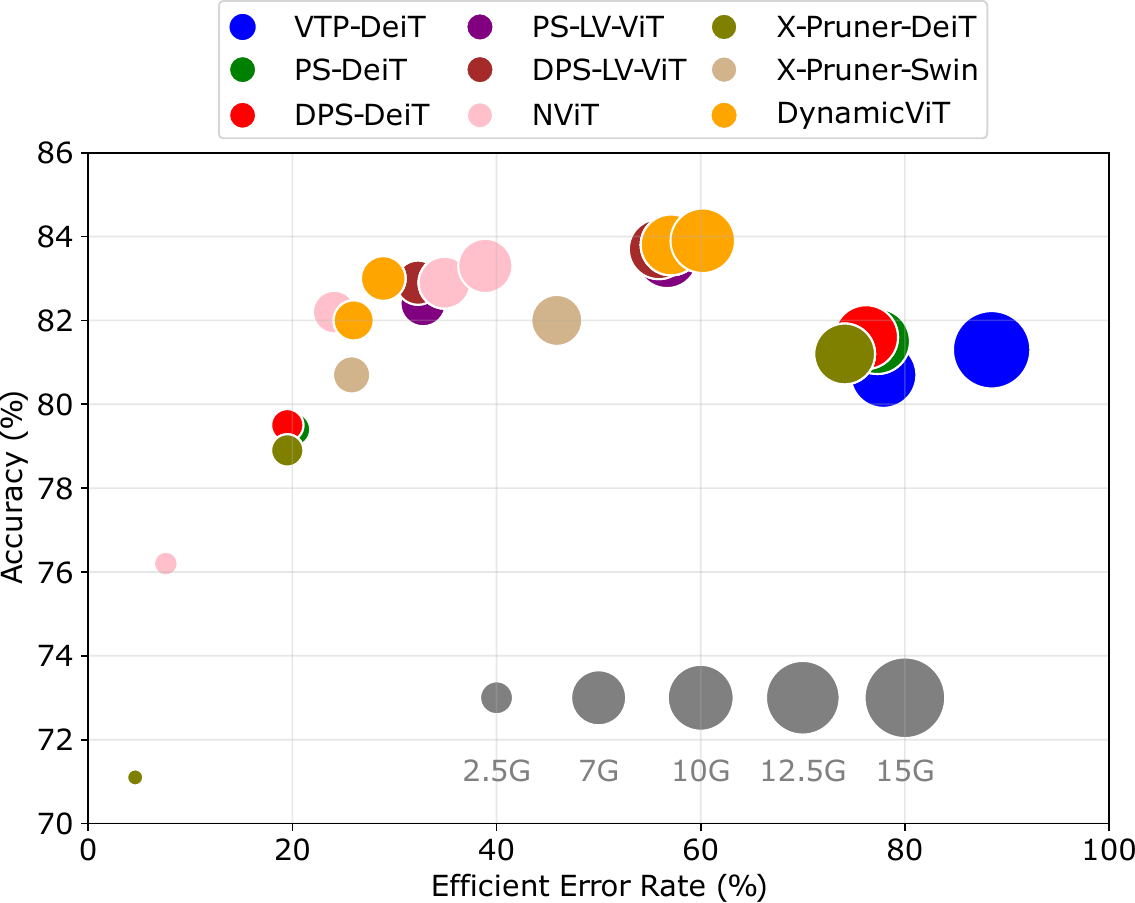}
    \caption{Graphical comparison of \textbf{P} model Top-1 accuracies and EER values on the ImageNet-1K dataset. The area of bubbles corresponds to the number of FLOPs. The reference FLOPs sizes (from 2.5G to 15G) are shown in gray in the bottom right corner.}
    \label{fig:P_models_results}
\end{figure}

\begin{figure}[h]
    \centering
    \includegraphics[width=0.95\linewidth]{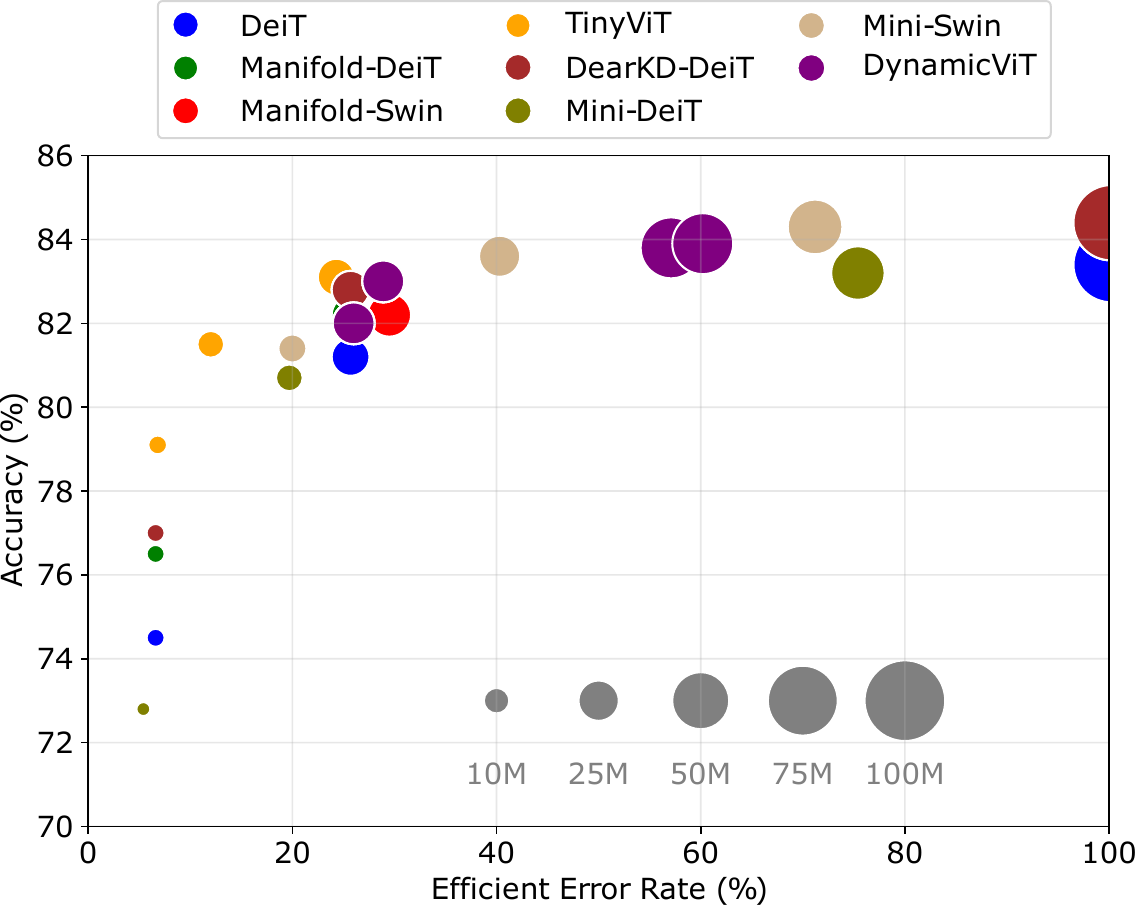}
    \caption{Graphical comparison of \textbf{KD} model Top-1 accuracies and EER values on the ImageNet-1K dataset. The area of bubbles corresponds to the number of trainable parameters (\#Param). The reference \#Param sizes (from 10M to 100M) are shown in gray in the bottom right corner.}
    \label{fig:KD_models_results}
\end{figure}

For a more in-depth investigation of the compared methodologies, Figures \ref{fig:CA_models_results}, \ref{fig:P_models_results}, and \ref{fig:KD_models_results} report a graphical representation of the model's distributions regarding accuracy-EER values for the CA, P, and KD efficiency categories, respectively.
The models with the best trade-off between accuracy and EER values are highlighted in orange.

\end{document}